\documentclass[serif,twocolumn, alpha-refs]{wiley-article}
\usepackage{siunitx}

\usepackage{textcomp}

\usepackage[hidelinks]{hyperref}
\hypersetup{
	colorlinks=true,
	linkcolor=black,
	filecolor=black,
	citecolor=black,
	urlcolor=blue,
}

\usepackage{booktabs}

\usepackage{multirow}

\UseRawInputEncoding
\usepackage{times}
\usepackage{epsfig}
\usepackage[ruled,vlined,linesnumbered]{algorithm2e}
\usepackage{pgfplots}
\pgfplotsset{compat=1.16}
\usepackage{comment}
\usepackage{xcolor}
\usepackage{lettrine}
\usepackage{lscape}
\usepackage{dirtytalk}
\usepackage{tabularx}
\usepackage{colortbl}
\usepackage{hhline}
\usepackage{longtable}
\usepackage{rotating}
\usepackage{pdflscape}
\usepackage{sidecap}
\usepackage[export]{adjustbox} % https://tex.stackexchange.com/questions/91566/syntax-similar-to-centering-for-right-and-left
\usepackage{acronym}
\acrodef{FCN}[FCN]{Fully Convolutional Network}
\acrodef{GAME}[GAME]{Grid Average Mean Absolute Error}
\acrodef{DL}[DL]{Deep Learning}
\acrodef{DNN}[DNN]{Deep Neural Network}
\acrodef{ML}[ML]{Machine Learning}
\acrodef{CV}[CV]{Computer Vision}
\acrodef{AI}[AI]{Artificial Intelligence}
\acrodef{CNN}[CNN]{Convolutional Neural Network}
\acrodef{JCU}[JCU]{James Cook University}
\acrodef{MAE}[MAE]{Mean Average Error}
\acrodef{MAP}[mAP]{Mean Average Precision}
\acrodef{CA}[CA]{Classification Accuracy}
\acrodef{LCFCN}[LCFCN]{Localisation-based Counting loss Fully Convolutional Network}
\acrodef{RUV}[RUV]{Remote Underwater Video}

\newcommand{\alz}[1]{\textcolor{black}{#1}}
\usepackage[most]{tcolorbox}
\newtcolorbox[auto counter]{pabox}[2][]{%
colback=green!5!white,colframe=green!75!black,fonttitle=\bfseries,
title=Box~\thetcbcounter: #2,#1}
\usepackage{graphicx}
\graphicspath{{./Graphics/}}
\DeclareGraphicsExtensions{.pdf,.jpeg,.png,.jpg}

\usepackage{amsmath}

\usepackage{array}

\usepackage{url}

\title{\alz{Computer Vision and Deep Learning for Fish Classification in Underwater Habitats: A Survey}}

\author[1]{Alzayat Saleh }
\author[1]{Marcus Sheaves }
\author[1,2]{Mostafa~Rahimi~Azghadi }

\affil[1]{College of Science and Engineering, James Cook University, Townsville, QLD, Australia}
\affil[2]{ARC Research Hub for Supercharging Tropical Aquaculture through Genetic Solutions, James Cook University, Townsville, QLD, Australia}

\corraddress{Mostafa Rahimi Azghadi, PhD, College of Science and Engineering, James Cook University, Townsville, QLD, Australia}
\corremail{mostafa.rahimiazghadi@jcu.edu.au}

\presentadd[]{College of Science and Engineering, James Cook University, Townsville, QLD, Australia}

\fundinginfo{This research is supported by an Australian Research Training Program (RTP) Scholarship, and the Australian Research Council funding through their Industrial Transformation Research Program.}

\runningauthor{Saleh et al.}

\begin{document}

\begin{frontmatter}
\maketitle

\begin{abstract}
Marine scientists use remote underwater image and video recording to survey fish species in their natural habitats. This helps them \alz{get a step closer toward understanding and predicting} how fish respond to climate change, habitat degradation, and fishing pressure. This information is essential for developing sustainable fisheries for human consumption, and for preserving the environment. However, the enormous volume of collected videos makes extracting useful information a daunting and time-consuming task for a human. %Smart automated \ac{CV} methods provide a solution for this. 
A promising method to address this problem is the cutting-edge Deep Learning (\ac{DL}) technology. %, which is a cutting-edge technology for image processing and data interpretation with great potential. 
 \ac{DL} can help marine scientists parse large volumes of video promptly and efficiently, unlocking niche information that cannot be obtained using conventional manual monitoring methods.
In this paper, \alz{we first provide a survey of Computer Visions (CV) and \ac{DL} studies conducted between 2003-2021 on fish classification in underwater habitats. We then give an overview of the key concepts of \ac{DL}, while analyzing and synthesizing DL studies.}
We also discuss the main challenges faced when developing \ac{DL} for underwater image processing and propose approaches to address them. Finally, we provide insights into the marine habitat monitoring research domain and shed light on what the future of \ac{DL} for underwater image processing may hold. \alz{This paper aims to inform marine scientists who would like to gain a high-level understanding of essential DL concepts and survey state-of-the-art DL-based fish classification in their underwater habitat.}
% This paper aims to inform marine scientists who would like to apply \ac{DL} in their research and learn the general concepts around deep learning, while surveying the prior art in a comprehensive literature. 
% to computer scientists who would like to survey state-of-the-art \ac{DL}-based underwater fish habitat monitoring literature.
%This paper aims to inform a wide range of readers, in particular marine scientists who would like to grasp a high-level understanding of \ac{DL} and see how it is evolving to facilitate their research efforts. At the same time, it is suitable for computer scientists who would like to survey state-of-the-art \ac{DL}-based methodologies for underwater habitat monitoring. 

% Please include a maximum of seven keywords
\keywords{Fish Habitat, Monitoring, Computer Vision, Deep Learning.}
\end{abstract}
\end{frontmatter}

\section*{Nomenclature}
	\begin{tabbing}
		AI ~~~~~~~~~ \= Artificial Intelligence\\
        ANN\>Artificial Neural Networks\\
        AUV \> Autonomous Underwater Vehicle\\
        CNN \> Convolutional Neural Network\\
        CV \> Computer Vision\\
        DL \> Deep Learning\\
        DNN \> Deep Neural Networks\\
        FCN\>Fully Convolutional Network\\
        LSTM\>Long short-term memory\\ 
        ML \> Machine Learning\\
        OCR \> Optical character recognition\\
        RNN \> Recurrent Neural Network\\
		ROV \> Remotely Operated Vehicles\\
        RUV \> Remote Underwater Video\\
	\end{tabbing}
	
\section{Introduction}\label{secintro}

% \IEEEPARstart{U}{nderstanding} 
Understanding and modelling how fish respond to climate change, habitat degradation, and fishing pressure are critical for environmental protection, and are  crucial steps toward ensuring sustainable natural fisheries, to support ever-growing human consumption  \citep{Zarco-Perello2019}. Effective monitoring is a vital first step  underpinning  decision support mechanisms for identifying problems and planning actions to preserve and restore the habitats. \alz{However, there is still a gap between the complexity of marine ecosystems and the available monitoring mechanisms.} 

Marine scientists use underwater cameras to record, model, and understand fish habitats and fish behaviour. \ac{RUV} recording in marine applications  \citep{Zarco-Perello2019} has shown great potential for fisheries, ecosystem management, and conservation programs  \citep{Piggott2020}.
With the introduction of consumer-grade high-definition cameras, it is now feasible to deploy a large number of \acp{RUV} or Autonomous Underwater Vehicles (AUVs) to collect substantial volumes of data and to perform more effective monitoring  \citep{Pope2010,Rasmussen2008,Thorstad2013}. However,
underwater  habitats  introduce diverse video monitoring challenges such as adverse water conditions, high similarity between fish species, cluttered backgrounds, and occlusions among fish. 
In addition, the volume of data generated by deployed RUVs and AUVs rapidly surpasses the capacity of human video viewers, making video analysis prohibitively expensive  \citep{Konovalov2019a}.
% Besides, traditional fish detection methods  \citep{Pope2010,Rasmussen2008,Thorstad2013} are usually unreliable in visually complex underwater ecosystems, therefore RUVs are required for conservation management  \citep{Pope2010}.
% These challenges render human-centred monitoring very difficult, if not impossible.
Moreover, humans are more prone to error than a well-designed machine-centred monitoring algorithm. 
%Effective monitoring can give managers information on which areas require preserving and restoration in order to maintain a healthy fish population for human use and environmental conservation. 
Therefore, an automated, comprehensive monitoring system could  significantly reduce  labour expenses while improving throughput and accuracy, increasing the precision in estimates of  fish stocks, fish distribution and  biodiversity in general  \citep{Hilborn1992}.
Implementing such  systems necessitates effective \acf{CV} processes. 
As a result, significant research has been conducted on implementing monitoring tools and techniques that build upon \ac{CV} algorithms for determining how fish exploit various maritime environments and differentiating between fish species  \citep{Zion2012}.

\begin{figure*}[!ht]
\includegraphics[width=\textwidth]{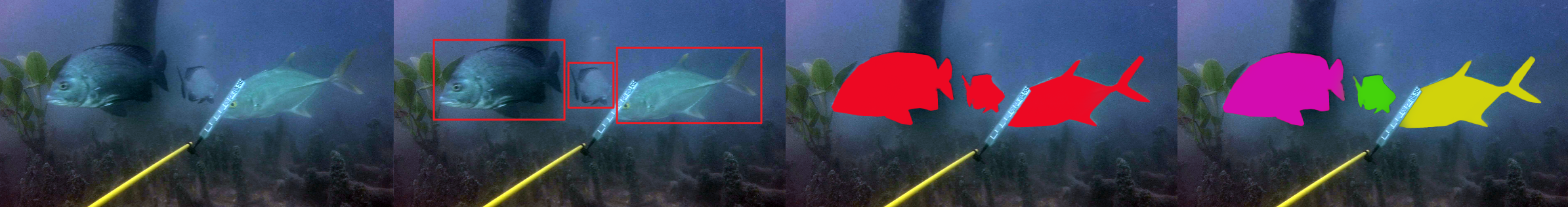}
\caption{\textbf{Illustration of four typical types of \ac{CV} tasks} From left: Image Classification (\textit{i.e.} is there a fish in the image, or what type (class) of fish is in the image?), Object Detection/Localisation, Semantic Segmentation,  Instance Segmentation.
} \label{CV_tasks}
\end{figure*}

% \input{tables/box1}

%An automated system for underwater fish habitats monitoring would address this challenge.

In image analysis and \ac{CV} domains, \acf{DL} approaches have consistently produced state-of-the-art results in a variety of applications from agriculture  \citep{Olsen2019DeepWeeds:Learning} to medicine  \citep{Saleh2021AImages,Azghadi2020HardwareApplications} using \acp{DNN}  \citep{Zheng2020,Miikkulainen2019EvolvingNetworks, Montavon2018MethodsNetworks}. 
\alz{Notably, a video is inherently composed of images or frames, which are processed using image analysis techniques. Therefore, image- and video-based monitoring tasks can be done using \ac{DL} models such as \acp{CNN} that receive an image (frame) as their input. Therefore, the methods mentioned for image-based tasks are useful for both images and videos.}

Many of \ac{DNN}-based approaches outperform conventional methods in marine applications, including ecological and habitat monitoring, using video trap data  \citep{Willi2019,Tabak2019}. 
\ac{DL} is a technique that mimics how people acquire knowledge by  continuous analysis of input data.
%\ac{DL} models  \citep{Liu2020} have made significant progress in discriminative tasks .
The main drivers of \ac{DNN} success over the past decade have been architectural progress by a large community of computer scientists, more powerful computers and processors, and access to massive amounts of data, which is critical for developing successful generalizable \ac{DL} applications.

\acp{DNN} have been successfully employed in many \ac{CV} applications such as object classification, identification, and segmentation as a result of the invention of \ac{CNN}.
\ac{CNN} is a class of \ac{DNN}, most commonly applied to visual analyses. For instance, \acp{CNN} have been successfully used for analysis of fish habitats  \citep{Xu2019,Konovalov2019a,Pope2010}. 
In comparison to other image recognition algorithms, \acp{CNN} have the significant benefit that they require limited pre-processing. \acp{CNN} are not hand-engineered but uncover and  learn hidden features in the data on their own. They  learn level-by-level with various levels of abstraction. 
For instance, they learn simple shapes (edges, lines, etc.) in the first few layers, understand more sophisticated patterns in their next layers, and learn classes of objects in their final layers. 
%In a \ac{CNN}, the image spatial features are maintained via parameterised, sparsely connected kernels. Convolutional layers then gradually reduce image spatial resolution while increasing the depth of their feature maps.
%This compilation of convolutional transformations can provide image representations that are far lower-dimensional and more useable than those produced by hand. 
%The success of \ac{CNN} has increased interest in and enthusiasm about using \ac{DL} to \ac{CV} challenges.

A putative challenge with \acp{CNN} is that they require a large number of images to be fully trained and generalise their learning to unseen scenarios. On the other hand, CNNs have an interesting and  powerful feature that enables transfer of their learning and knowledge across different domains.  This means that they can be fine-tuned to work on new datasets (e.g. fish datasets) other than the one that they have been trained on (e.g. general objects). \alz{However, fine-tuning with annotated datasets specific for a given domain implies cost/effort/time needed to generate the annotations, and also requires a larger set of data which may not always be available.} %such as ImageNet  \citep{Li2009}  and COCO  \citep{Lin2014}.
Equipping  \ac{CV} algorithms with the powerful learning and inference capabilities of \acp{CNN} can provide marine scientists and ecologists with powerful tools to help them better understand and manage marine environments. However, although \ac{DL}, and its variants such as \acp{CNN}, have been applied to various applications across a multitude of domains  \citep{Deng2013,Pathak2018,Min2017}, their use in conjunction with computer vision for marine science and fish habitat monitoring is not broadly appreciated, meaning they remain under utilised. To address this, in this paper, we introduce key concepts and typical architectures of \ac{DL}, and provide a comprehensive survey of key \ac{CV} techniques for underwater fish habitat monitoring. % including classification, counting, localisation, and segmentation. 
In addition, %we survey of publicly available underwater fish datasets, and compare various \ac{DL} techniques in the underwater fish habitat monitoring domain, to
we provide insights into challenges and opportunities in the underwater fish habitat monitoring domain.
% \alz{In summary, our article is an introduction to deep learning for marine scientists, with follow-up discussions on the use of deep learning and the general concepts introduced in the marine task of underwater fish classification.}
\alz{It is worth noting that our article is written to provide a general and high-level, as opposed to detailed, introduction of deep learning and its relevant contexts for marine scientists. This is useful in understanding the  follow-up discussions on the use of deep learning in the marine task of underwater fish classification.}

% Although a recent survey reviews deep learning techniques for marine
% ecology \cite{Goodwin2021} and briefly discusses DL-based fish image analysis, to the best of our knowledge, no comprehensive survey and overview of deep learning with a specific focus on \alz{fish classification in underwater habitats} currently exist. Our paper tries to address this gap and to facilitate the application of modern deep learning approaches into the challenging underwater fish images analysis and monitoring domains. We do this by comprehensively review and analyse the literature providing information, in on the DL model they have used, their training dataset, their annotation techniques, their performance and a comparison to other similar works. This detailed analysis is not provided in [22]. 
% \alz{Unlike \cite{Goodwin2021} and  \cite{Li2021a}, our article comprehensively reviews and analyses the literature providing information on the DL model they have used, their training dataset, their annotation techniques, their performance, and a comparison to other similar works.}

\alz{Although a recent survey reviews deep learning techniques for marine ecology  \citep{Goodwin2021} and briefly discusses DL-based fish image analysis, to the best of our knowledge, no comprehensive survey and overview of deep learning with a specific focus on fish classification in underwater habitats currently exists. Our paper tries to address this gap and to facilitate the application of modern deep learning approaches into the challenging underwater fish images analysis and monitoring domains. We do this by comprehensively reviewing and analysing the literature providing information about the DL model the previous works have used, their training dataset, their annotation techniques, their performance and a comparison to other similar works. This detailed analysis is not provided in  \citep{Goodwin2021}.} 

\alz{In addition, another survey  \citep{Li2021a} exists that focuses on five different tasks of classification, detection, counting, behaviour recognition, and biomass estimation. Compared to  \citep{Li2021a}, we provide a different analysis and review of the literature because we mainly focus on the classification of fish in underwater images. Li and Du's work  \citep{Li2021a} fits mostly in the domain of aquaculture, while our paper is mostly a review of "fish classification techniques in underwater habitats" and the challenges they bring. Li and Du introduce a background to many different DL architectures, one of which is CNN, which is the focus of our paper. Also, the challenges and opportunities that Li and Du introduce are different to our paper, which is mainly about underwater fish classification in their natural habitat.} 

\alz{Furthermore, we provide a historical review of the CV and DL research using underwater cameras for fish classification, and analyse how their accuracy has evolved over years. This is not covered by previous works including \citep{Goodwin2021, Li2021a}. 
}

\section{Background To Computer Vision and Machine Learning}\label{seccv}
 
Humans, have a natural ability to comprehend the three-dimensional structure of the world around us.
Vision scientists  \citep{Oomes2001} have spent decades attempting to understand how the human visual system functions  \citep{Wang2017}.
Inspired by their findings, \ac{CV} researchers  \citep{Ballard1982ComputerVision,Huang1996ComputerPromise,Sonka2008ImageVision} have also been working on ways to recover the 3D shape and appearance of objects from photos.
The automatic retrieval, interpretation, and comprehension of useful information from a single image or collection of images can be referred to as \ac{CV}. In another definition, \ac{CV} is a field of \ac{AI} that focuses on training computers to detect, recognise, and understand images similarly to processes used by  humans. This necessitates the development of logical and algorithmic foundations for automated visual understanding   \citep{Mader2018}. This understanding can include image classification, object localisation, object recognition, semantic segmentation, and instance segmentation, as shown in Figure \ref{CV_tasks}. Today, computers with \ac{CV} powers can extract, analyse, and interpret significant information from a single image or a sequence of images.

%\ac{CV}  \citep{Ballard1982ComputerVision,Huang1996ComputerPromise,Sonka2008ImageVision} is an interdisciplinary research field that studies how computers may be used to extract high-level understanding from visual pictures or videos.
Despite this progress, the goal of making a computer to understand a picture at the same level as a two-year-old child remains unattainable. 
This is due, in part, to the fact that \ac{CV} is an inverse problem in which we attempt to recover specific unknowns despite having inadequate knowledge to completely describe the solution.
In \ac{CV} applications, the cause is usually an exploration process, while the effects are the observed data. The corresponding forward problems then consist of predicting empirical data given complete knowledge of the exploration process. In some sense, solving inverse problems means \say{computing backwards}, which is usually more difficult than forward problem solving  \citep{Hohage2020}.
 
% From the standpoint of computer science, \ac{CV} seeks to simplify functions analogous to the human visual system.
%The automatic retrieval, interpretation, and comprehension of useful information in a single image or collection of images is referred to as \ac{CV}.
%It necessitates the development of a logical and algorithmic foundation for automated visual understanding  \cite{Mader2018}. 
%Other topics covered by  \ac{CV} and \ac{AI} \cite{Shapiro1992Encyclopedia1,Ligeza1995,Sundvall2019}  include pattern recognition and learning techniques.
%As a result, \ac{CV} can also be used as part of 
%the discipline of \ac{AI} or the field of computer science in general.
%\ac{CV} involves the methodology underlying artificial systems that extract information from images. 
%'Images' in \ac{CV} are obtained from several sources, including video streams, collections of camera pictures, and multidimensional images from medical devices \cite{frouin1992famis}. As a scientific field, \ac{CV} seeks to apply its concepts and principles to the development of \ac{CV} applications.
%\ac{CV} focuses on numerous distinct tasks, including image classification, object localisation, object recognition, semantic segmentation, and instance segmentation (see Figure \ref{CV_tasks}).
 
%Many challenges that were previously difficult for traditional software development tools and techniques were solved thanks to \ac{ML}.

\begin{figure}[!ht]
\includegraphics[width=0.48\textwidth]{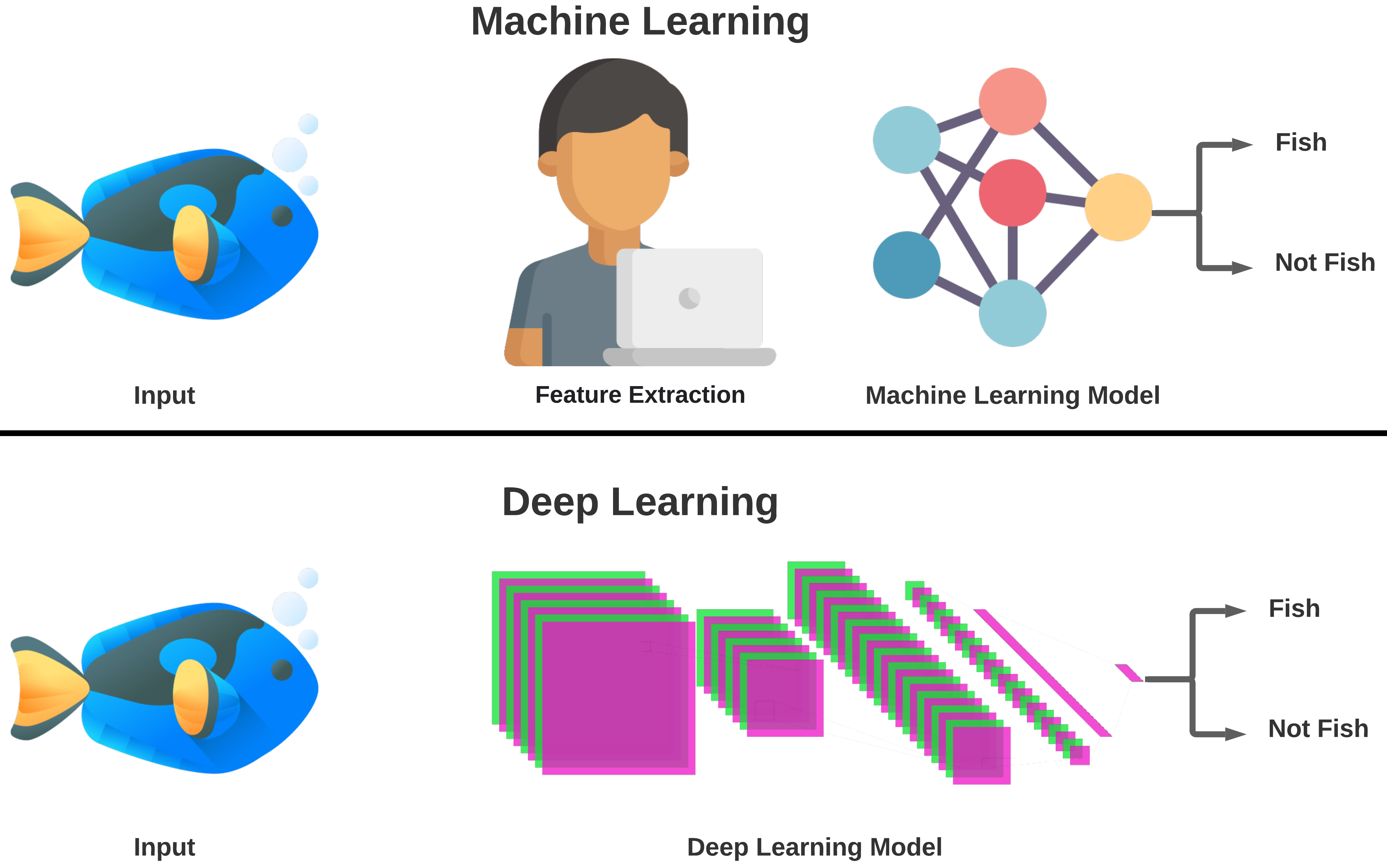}
\caption{\textbf{Comparison between \ac{ML} and \ac{DL}.}
In \ac{ML} techniques, the features need to be extracted by domain expert while \ac{DL} relies on layers of artificial neural networks to extract these features.
} \label{CV_comp}
\end{figure}

The problem of backward computation was eased by the introduction of \ac{ML} techniques more than 6 decades ago. However, in conventional \ac{ML} approaches, the majority of complex features of the learning subject must be identified by a domain expert in order to decrease the complexity of the data and make patterns more evident for successful learning (see Figure \ref{CV_comp}-top).
However, \ac{DL} offered a fundamentally new method to \ac{ML}. Most \ac{DL} algorithms possess the ground-breaking ability of automatically learning high-level features from data with minimal or no human intervention (see Figure \ref{CV_comp}-bottom).

\ac{DL} is based on neural networks, which are general-purpose functions that can learn almost any data type that can be represented by many instances. 
When you feed a neural network a large number of labelled instances of a certain type of data, it will be able to uncover common patterns between those examples and turn them into a mathematical equation that will assist in categorising future data.
Empowered by this fundamental feature, \ac{DL} and \ac{DNN} have progressed from theory to practice as a result of advancements in hardware and cloud computing resources  \citep{Azghadi2020HardwareApplications}.
In recent years, \ac{DL} approaches have outperformed previous state-of-the-art \ac{ML} techniques in a variety of areas, with \ac{CV} being one of the most notable examples.

Before the introduction of \ac{DL}, the capabilities of \ac{CV}  were severely limited, necessitating a great deal of manual coding and effort. 
However, owing to improved research in \ac{DL} and neural networks,  \ac{CV} is now able to outperform humans in several tasks related to object recognition and classification \citep{Sarigul2017c, Salman2016, Qin2016a, Sun2017a}. %A clear example of this progress is shown in Table \ref{table:imagenet}, where  that shows the enhanced performance of state-of-the-art models, using extra data from Instragram \citep{Mahajan2018} and JFT  \citep{Hinton2015}, on the challenging ImageNet classification task under  the Top-1 (meaning the model prediction is exactly correct) and Top-5 (meaning one of the top five predictions made by the model, is correct) criteria. 
% accuracy of state-of-the-art models on the challenging ImageNet classification task is shown. The table shows that newer models have used extra training data from  Instagram \citep{Mahajan2018} and  JFT  \citep{Hinton2015} to improve classification performance even more.
% \ac{CV} is being used today in a wide variety of real-world applications, which include:
 \ac{CV} equipped with \ac{DL}, is being used today in a wide variety of real-world applications, that include, but are not limited to: %as shown below: %in Box~\ref{box4}.

\begin{itemize}

\item \textit{Optical character recognition (OCR)}  \citep{Converso1990}: automatic number plate recognition and reading handwritten postal codes on letters;

\item \textit{Machine inspection}  \citep{Park2016}: fast quality assurance inspection of components using stereo vision with advanced lighting to assess tolerance levels on aircraft wings or car body parts, or to spot flaws in steel castings using X-ray technology;

\item \textit{Retail}  \citep{Trinh2012}: object detection for automatic checkout lanes;

\item \textit{Medical imaging}  \citep{Erickson2017}: registration of preoperative and intra-operative imaging or long-term analyses of human brain anatomy as they age;

\item \textit{Automotive safety}  \citep{Falcini2017}: detection of unforeseen objects such as pedestrians on the street (e.g. fully autonomously driving vehicles);

\item \textit{Surveillance}  \citep{Brunetti2018}: Monitoring of trespassers, studies of highway traffic, and monitoring pools for drowning victims;
\item \textit{Fingerprint recognition and bio-metrics}  \citep{Kim2016}: For both automatic entry authentication and forensic software.
\end{itemize}

This demonstrates  the significant impact of \ac{DL} on CV and demonstrates its potential for marine visual analysis applications.
% The following section provides an overview of some of the most commonly utilised \ac{DL} schemes in \ac{CV}.

\begin{figure*}[!t]
\includegraphics[width=0.99\textwidth]{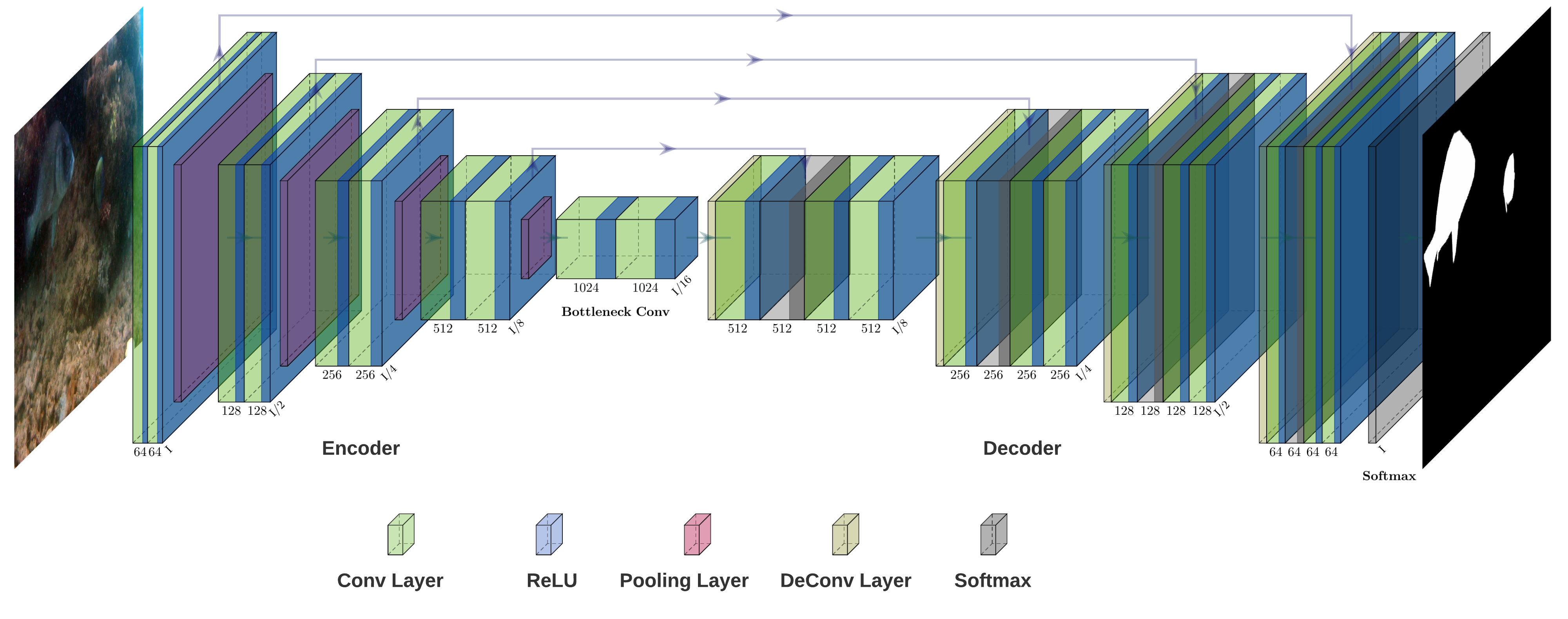}
\caption{A popular \ac{CNN} architecture, named UNET  \citep{Ronneberger2015} is demonstrated. The first component of UNET is the encoder, which is used to extract features from the input image. The second component is the decoder that outputs per-pixel scores. The network is composed of five different layers including convolutional (Conv Layer), Rectified Linear Unit (ReLU), Pooling, Deconvolutional (DeConv), and Softmax.%Here, the white blobs are predicted segmentation of the fish body. 
Here, the task of the DNN layers has been to give a high score to only the pixels in the input image that belong to the fish body, resulting in the demonstrated white blobs output, showing where the fish are.}
% a classic \ac{CNN} for digits recognition. Each layer is a feature map, i.e a set of different neural convolutions, whose weights are constrained to be identical.}
\label{fig:VGG}
\end{figure*} 

\alz{
\section{The evolution of Computer vision approaches to fish classification}\label{sectrend}
%The advent of consumer-grade cameras, the exponential increase in computational power and the availability of large image and video datasets have led to major advances in the computer vision for fish classification. In fact, the last two decades have witnessed the emergence of novel computer vision approaches for fish classification including the design and evaluation of complex algorithms that could not be applied before given the availability of sufficiently large data and the use of powerful GPUs. Therefore, we carried out a systematic literature review over the last two decades and reported the results in this section.
The last two decades have witnessed the emergence of novel computer vision approaches for fish classification including the design and evaluation of complex algorithms that could not be applied before and became possible with the availability of sufficiently large data and the use of powerful Graphical Processing Units (GPUs). Here, we perform a systematic literature review of the evolution of computer vision applications and their different approaches over the past two decades.
} 

\alz{
\subsection{Search and Selection Criteria}
We systematically reviewed the literature for underwater fish classification using computer vision  from 2003 to 2021. The search terms used included
"underwater fish classification", "Deep Learning", "Computer Vision", "Machine vision".
The databases searched included Wiley Online Library, IEEE Xplore, Elsevier/ScienceDirect, and ACM Digital Library. We believe that combining these four databases accurately represents global research on this topic.}

\begin{figure}[!t]
\centering
\includegraphics[width=0.49\textwidth]{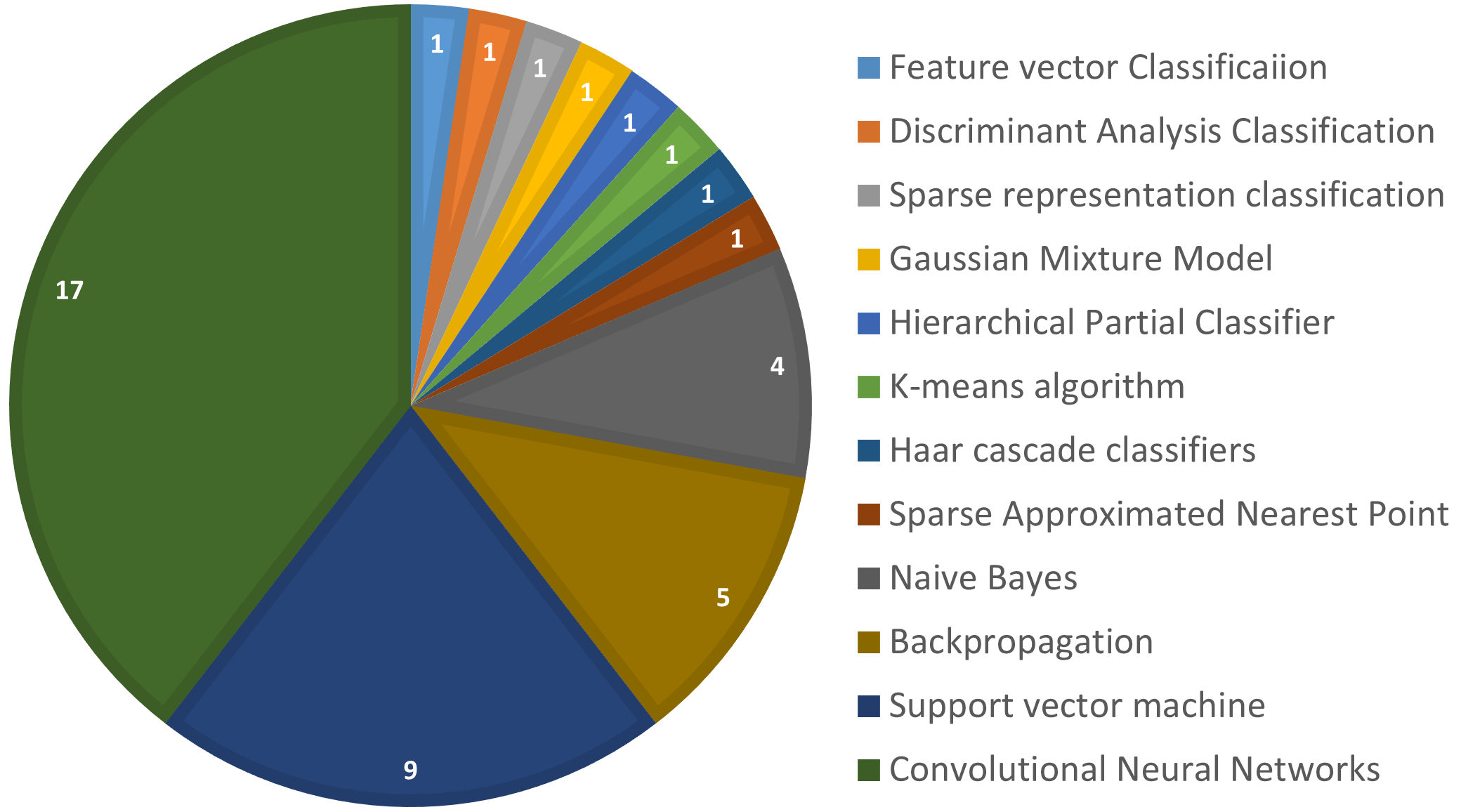}
\caption{\alz{An overview of the methods used for fish classification using different Computer Vision techniques from 2003 to 2021. It is evident from the graph that DL and its CNNs have attracted more attention than classical ML methods.}} %See tables \ref{table:cls1} to \ref{table:trend} for a more detailed comparison.}}
\label{fig:trend}
\end{figure}

\alz{
We divided the search into two stages. First, we queried the databases for articles with the above-mentioned keywords in their titles and contents.
Secondly, we independently reviewed the titles and abstracts of each article in order to check its relevance to our research topic. 
After the individual title and abstract reviews, we considered 64 articles for full-text reading. In the full-reading phase, we extracted information relevant to our research topic. In this phase, it became clear that 21 papers were not relevant to our work and therefore were excluded. This left us with 43 papers for fish classification, 26 of which were classical Computer Vision methods, and 17 Deep Learning papers. Figure \ref{fig:trend} presents an overview of the methods used in the identified studies and classifies them into several groups, based on their classification algorithms that can be categorized into two general category of conventional CV, and modern DL models.}

\begin{figure*}[!h!t]
\centering
\includegraphics[width=0.99\textwidth]{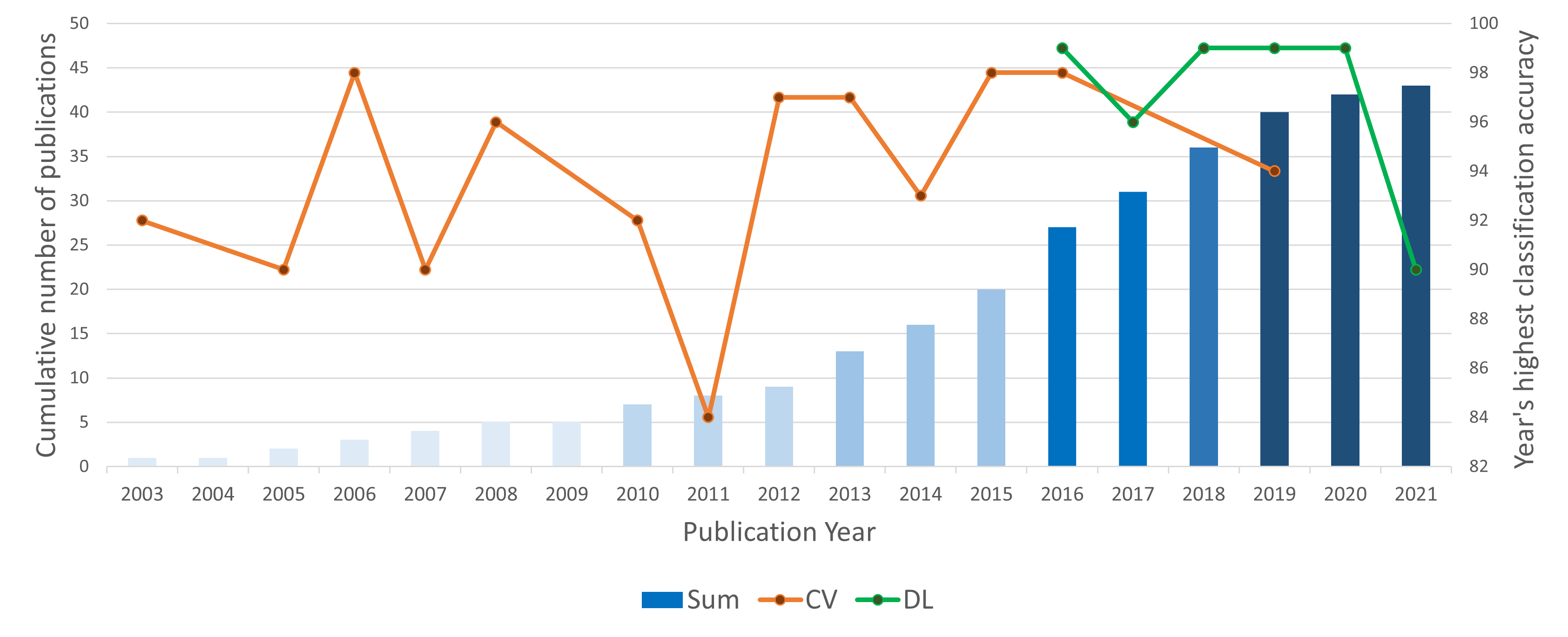}
\caption{\alz{An overview of the publication trend and performance of an extensive range of fish classification Computer Vision (CV) and Deep
Learning (DL) models from 2003 to 2021. Here the bars show the cumulative number of publications over years and the growth thereof, while the line graphs demonstrate the highest classification accuracy in each year in literature on the right-hand-side vertical axis.}}
\label{fig:accuracy}
\end{figure*}

\alz{
\subsection{The Evolution of Fish Classification Algorithms over Two Decades}
\textbf{The publication trend} for fish classification studies is summarized in Fig.~\ref{fig:accuracy}. The figure shows the cumulative number of publications and how the studies evolved over the past two decades. It is evident that the number of publications has been gradually increasing, but in 2016, when the first few studies using deep learning were combined with CV methods, the study numbers have seen the highest increase and a fast upward trajectory for a few years (2015-2019) after DL burgeoned in fish classification, and before slowing down. }

\alz{
Fig.~\ref{fig:accuracy} also shows the highest classification accuracy achieved in each year, as \textbf{a quality assessment metric}. It is evident that since 2016, when DL techniques were first proposed for fish classification, the accuracy has seen its highest value. At the same time, it can be seen that there are large differences in the accuracies achieved over years. The main reasons for this difference include (i) using different classification and CV methods, and (ii) using different fish image sources that were captured differently and in different environments. These bring huge variations among studies, such as different image resolutions and inconsistent resolutions and image qualities across time. For example, some fish image datasets are in grayscale  \citep{Chuang2014, Chuang2016, Kartika2017}, while others are in colour  \citep{Zion2008, Zion2007, Shafait2016}. Some datasets contain only images  \citep{Islam2019, Kartika2017}, while others include videos  \citep{Lopez-Villa2015, Cutter2015, Hossain2016}. Also, some datasets  \citep{Huang2014} used low-quality images from the internet, which negatively affects the accuracy, due to their wide range of resolutions, colours, and angles. They are also taken at random locations. Due to these factors in various studies, direct comparison of accuracy values is unfeasible, though the accuracy trend can be still observed in Fig. \ref{fig:accuracy}.}

%It is worth noting that, since various methods and fish datasets were used in these studies,  }

%\alz{
%\textbf{To assess the quality criteria} in the different studies over time, we compared the classification accuracy of each study. It has been seen that there are large differences in the classification accuracy of each individual study. There are several reasons for the variations: different methods and different image sources. Some studies use an open dataset and some others use a fish dataset created by themselves. }

%\alz{
%The quality criteria of the dataset affect the accuracy of the classification. Because different datasets have different resolutions, the resolutions and image qualities are not consistent across time and studies. For example, some datasets are provided in grayscale, while some of them are provided in colour. Some datasets contain only images, while others include videos. Therefore, the differences in datasets from different times have led to different accuracies in classification. As for the source of the data, we have found that using images from the internet also leads to low-quality datasets, which negatively affects the accuracy. This phenomenon can be explained by the fact that these datasets have a wide range of resolutions, colours, and angles, and are taken at random locations. These images are not easy to capture or clean. }

\alz{
Computer vision for fish classification in the early 2000s and up to 2016, when first DL works started, has been mainly to manually extract fish features and then build classifiers that recognize these features. These conventional studies are listed, in a chronological order, in Table~\ref{table:trend}. 
Although there are many existing models, most of the classical non-DL models are based on local and engineered features. These include works using Haar features \citep{Mutneja2021}, Scale-Invariant Feature Transform (SIFT) \citep{Lindeberg2012}, and Histogram of Oriented Gradient (HOG) \citep{Dalal2005}, which need hand-engineered algorithms. Because these algorithms are not suitable for recognizing images of untrained animals and cannot capture fish features from complex backgrounds, they usually use a large number of manually extracted samples to build classifiers.}

\alz{
As shown in Table \ref{table:trend}, support vector machines  \citep{Rova2007,Hu2012b,Fouad2014,Huang2014,Chuang2016,Ogunlana2015,Hossain2016, Wang2017a, Islam2019} were one of the most commonly used classifiers for fish recognition, but they are prone to overfitting when trained with too many samples. This problem limits the scale of application. Another popular classification technique used in early works was backpropagation to train a simple feed-forward shallow neural network  \citep{Alsmadi2010,Alsmadi2011,Pornpanomchai2013,Badawi2014, Boudhane2016a}. Although this technique can handle simple samples, it is difficult to scale because of the neural network shallow layers, which will be explained in the next Section. Naive Bayes  \citep{Nery2005,Zion2007,Zion2008,Kartika2017} have also been used to classify fish since the early 2000s and up to 2017. The technique does not require much training data, and as shown in Table \ref{table:trend} can reach good accuracy levels. Table \ref{table:trend} also shows some other CV classification techniques, which while not as popular as the above-mentioned methods, could demonstrate good performance. However, it should be noted that, most of the CV techniques in Table \ref{table:trend}, were carefully engineered for their target datasets and are not capable of showing a similar performance level if used for another similar dataset. They will perhaps require an overhaul in their design, starting from manual feature engineering, to designing the detailed classification models. 
}

\alz{
In contrast, deep learning can extract features and perform classification tasks automatically. The features are invariant to data scaling, translation, rotation, and distortion. Because these features are better for classification, the classification performance can be better than that conventional CV tasks using manually designed features. Also, DL classification models, compared to traditional CV one, usually require a simpler redesign procedure to work on a new similar dataset, due to the ability to extract features on their own.}

\alz{ 
Although DL emerged in 2012 \citep{ImageNet2012}, its first use for underwater fish classification was in 2016  \citep{Salman2016}. After that, 16 other works also used DL and its \acp{CNN}, as shown in Fig. \ref{fig:trend}, to develop models that learn features from large amounts of data without manual interference. These studies have shown that, by using deep learning, some of the usual fish image classification challenges such as image noise reduction, classification of difficult or rare-seen fish, and classifying small fish, can be solved.} 

\alz{In the following parts of this paper, we mainly focus on deep learning, how it works, and how it can be applied to develop efficient and high-performance underwater fish classifiers. We will also critically analyse the 17 DL studies found as part of our systematic literature review described earlier.}

% \usepackage{array}
% \usepackage{ragged2e}
% \usepackage{graphicx}
% \usepackage{colortbl}

%\appendix

%\section{Appendix A}
%\label{app:trend}

\begin{table*}
% \color{blue}
\caption{A list of computer vision studies for underwater fish classification between 2003-2021 using conventional classifiers and based on engineered features. The last column presents the work's achieved accuracy.}
\label{table:trend}
\centering
\resizebox{\linewidth}{!}{%
\begin{tabular}{>{\hspace{0pt}}m{0.585\linewidth}>{\RaggedLeft\hspace{0pt}}m{0.042\linewidth}>{\hspace{0pt}}m{0.273\linewidth}>{\RaggedLeft\hspace{0pt}}m{0.036\linewidth}}
\rowcolor[rgb]{0.663,0.655,0.655} \textbf{Article} & \multicolumn{1}{>{\hspace{0pt}}m{0.042\linewidth}}{\textbf{Year}} & \textbf{Classification Method} & \multicolumn{1}{>{\hspace{0pt}}m{0.036\linewidth}}{\textbf{AC}} \\
An Automated Fish Species Classification and Migration Monitoring System \citep{Lee2003} & 2003 & Feature vector Classification & 92 \\

Determining the appropriate feature set for fish classification tasks \citep{Nery2005} & 2005 & Naive Bayes & 90 \\

Real-time underwater sorting of edible fish species \citep{Zion2007} & 2006 & Naive Bayes & 98 \\
One Fish, Two Fish, Butterfish,  Trumpeter: Recognizing Fish in Underwater Video \citep{Rova2007} & 2007 & Support vector  machine & 90 \\

Classification of guppies' (Poecilia reticulata) gender  by computer vision \citep{Zion2008} & 2008 & Naive Bayes & 96 \\

Automatic Fish Classification for  Underwater Species Behavior Understanding \citep{Spampinato2010} & 2010 & Discriminant Analysis Classification & 92 \\

Fish Recognition Based on Robust Features  Extraction from Size and Shape Measurements Using Neural Network \citep{Alsmadi2010} & 2010 & Backpropagation & 86 \\

Fish Classification Based on Robust  Features Extraction From Color Signature Using Back-Propagation Classifier \citep{Alsmadi2011} & 2011 & Backpropagation & 84 \\

Fish species classification by color, texture and multi-class support vector  machine using computer vision \citep{Hu2012b} & 2012 & Support vector  machine & 97 \\

Real-world underwater fish recognition and identification, using  sparse representation \citep{Hsiao2014} & 2013 & Sparse representation classification & 81 \\

A research tool for long-term and continuous analysis of fish assemblage in coral-reefs using underwater camera footage \citep{Boom2014} & 2013 & Gaussian Mixture Model & 97 \\

Automatic Nile Tilapia Fish  Classification Approach using Machine Learning Techniques \citep{Fouad2014} & 2013 & Support vector  machine & 94 \\

Shape- and Texture-Based Fish Image  Recognition System \citep{Pornpanomchai2013} & 2013 & Backpropagation & 90 \\

A General Fish Classification Methodology  Using Meta-heuristic Algorithm With Back Propagation Classifier \citep{Badawi2014} & 2014 & Backpropagation & 80 \\

GMM improves the reject option in  hierarchical classification for fish recognition \citep{Huang2014} & 2014 & Support vector  machine & 74 \\

Supervised and Unsupervised Feature  Extraction Methods for Underwater Fish Species Recognition \citep{Chuang2014} & 2014 & Hierarchical Partial Classifier & 93 \\

A Feature Learning and Object Recognition  Framework for Underwater Fish Images \citep{Chuang2016} & 2015 & Support vector  machine & 98 \\

A novel tool for ground truth data  generation for video-based object classification \citep{Lopez-Villa2015} & 2015 & K-means algorithm & 93 \\

Automated detection of rockfish in  unconstrained underwater videos using Haar cascades and a new image dataset:  labeled fishes in the wild \citep{Cutter2015} & 2015 & Haar cascade classifiers & 89 \\

Fish Classification Using Support Vector  Machine \citep{Ogunlana2015} & 2015 & Support vector  machine & 79 \\

Fish identification from videos captured  in uncontrolled underwater environments \citep{Shafait2016} & 2016 & Sparse Approximated Nearest Point  & 94 \\

Fish Activity Tracking and Species  Identification in Underwater Video \citep{Hossain2016} & 2016 & Support vector  machine & 91 \\

Koi Fish Classification based on HSV  Color Space \citep{Kartika2017} & 2016 & Naive Bayes & 97 \\

Optical Fish Classification Using  Statistics of Parts \citep{Boudhane2016a} & 2016 & Backpropagation & 95 \\

Shrinking Encoding with Two-Level  Codebook Learning for Fine-Grained Fish Recognition \citep{Wang2017a} & 2017 & Support vector  machine & 98 \\

Indigenous Fish Classification of  Bangladesh using Hybrid Features with SVM Classifier \citep{Islam2019} & 2019 & Support vector  machine & 94\\

\end{tabular}
}
\end{table*}

\section{Background To Deep Learning}\label{secdl}

% \ac{DL}  \citep{goodfellow2016deep} is a sub-field of \ac{ML} that allows machines to learn through experience and to understand the universe in terms of a hierarchy of concepts \cite{Gao2019}. When a computer gathers information through experience, there is no need for a computer user to formally define all the knowledge that a computer requires \cite{YannLeCunYoshuaBengio2015}. The hierarchy of concepts helps a computer to learn abstract concepts from perception.
% \ac{DL} allows neural networks consisting of several layers 
% to learn data representation at multiple abstraction stages. 
% These approaches have significantly enhanced the field in terms of state-of-the-art voice recognition, visual object recognition, object identification and other tasks including drug discovery and genomics research. 
% \ac{DL} by so-called deep \ac{CNN} uses an subtle framework in large amounts of data by using a backpropagation algorithm \cite{Rojas1996} to determine how the computer can adjust the internal parameters that have been used to determine the representation in each layer from the representation in the preceding layer. 
% Deep convolutional networks have contributed to advancements in image, video, speech, and audio recognition, 
% while recurrent networks have shed light on recurrent data such as text and speech.

\acf{DL}  \citep{goodfellow2016deep,YannLeCunYoshuaBengio2015} is a subset of \ac{ML} algorithms that employs a neural network with several layers to very loosely replicate the function of the human brain by enabling it to "learn" from huge quantities of data. The learning happens when the neural network extracts higher-level features from input training data. The term "deep" refers to the usage of several layers in the neural network.
Lower layers, for example in image processing, could detect edges, whereas higher layers might identify parts of the object.
%In \ac{DL}, the term "deep" refers to the usage of several layers in the neural network.
%These neural networks seek to replicate the function of the human brain by enabling it to "learn" from huge quantities of data, however, they fall well short of its capabilities compared to human brain. 
%While a single-layer neural network may generate approximate predictions, more hidden layers can assist to optimise and improve for accuracy.
%\ac{DL} methods have grown in popularity as a result of their capacity to auto-detect features in complicated, high-dimensional data with high prediction accuracy.

\subsection{How Deep Learning differs from Machine Learning}
\acf{ML} is usually referred to as a class of algorithms that can recognise patterns in data and create prediction models automatically.
\acf{DL} is a subclass of standard \ac{ML} because it uses the same type of data and learning methods that \ac{ML} applies. 
However, when dealing with unstructured data, e.g. text and images, \ac{ML} usually goes through some pre-processing to convert it to a structured format for learning.
%This does not mean it never utilises unstructured data; it just means that it usually goes through some pre-processing to convert it to a structured format.
\ac{DL}, on the other hand, does not usually require the data pre-processing needed by \ac{ML}. It is capable of recognising and analysing unstructured data, as well as automating feature extraction, significantly reducing the need for human knowledge (see Figure \ref{CV_comp}-bottom).

For example, to recognise fish in an image, \ac{ML} requires that specific fish features (such as shape, colour, size, and patterns) be explicitly defined in terms of pixel patterns.
This may be a challenge for non-\ac{ML} specialists because it typically requires a deep grasp of the domain knowledge and good programming skills.
\ac{DL} techniques, on the other hand, skip this step entirely.
Using general learning techniques, \ac{DL} systems can automatically recognise and extract features from data.
This means that we just need to tell a \ac{DL} algorithm whether a fish is present in an image, and it will be able to figure out what a fish looks like given enough examples.
Decomposing the data into layers with varying levels of abstraction enables the algorithm to learn complex traits defining the data, allowing for an automatic learning approach.
\ac{DL} algorithms may be able to determine which features (such as fishtail) are most important in differentiating one animal from another.
Prior to \ac{DL}, this feature hierarchy needed to be determined and created by hand by an \ac{ML} expert.

\subsection{How Deep Learning works}
\acf{DNN}, also known as artificial neural network, is the basis of deep learning. \acp{DNN} use a mix of data inputs, weights, and biases to learn the data, by properly detecting, categorising, and characterising objects in a given dataset of interest.
\acp{DNN} are made up of several layers of linked nodes, each of which improves and refines the network prediction or categorisation capabilities.  For instance, Fig. \ref{fig:VGG} shows a popular \ac{DNN} architecture for image processing, called UNET  \citep{Ronneberger2015}. UNET, which is a fairly complex deep learning architecture, is composed of a few different components and layers, to achieve a specific learning goal, i.e. to segment fish body in an input image. 

\begin{figure}[!t]
\centering
\includegraphics[width=0.48\textwidth]{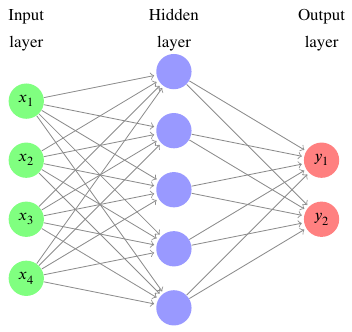}
\caption{A diagram of a single-layer neural network, composed of input, hidden, and output layers.}
\label{fig:nn}
\end{figure}

Any \ac{DNN} is composed of three types of layers, namely input, output, and hidden layers.
The visible layers are the input and output layers (see Figure \ref{fig:nn}). The \ac{DL} model gets the data for processing in the input layer, and the final prediction or classification is generated in the output layer. In a typical neural network, including a \ac{DNN}, the learning happens through two general processes, \textit{i.e.} forward and backward propagations. Forward propagation refers to the propagation of input data through the network layers to generate a prediction or classification result.
Backward propagation or, backpropagation in short, is where the learning happens in the network. Backpropagation uses a training model that %employs techniques such as gradient descent to 
determines prediction errors and then changes the weights and biases of the neural network by going backwards through its layers. 
Forward propagation and backpropagation work together to allow a neural network to generate predictions and reduce the network errors.
% (see Figure \ref{fig:bb}). 
Through many iterations of backward and forward propagation, the neural network prediction or classification accuracy improves.
%\ac{DNN} has a hierarchical composition that allows them to learn data representations with several levels of abstraction, the ability to learn highly complicated functions, and the ability to learn feature representations directly and automatically from data with minimum domain expertise.

Almost all \acp{DNN} work on and through the same principles described above. However, different \ac{DL} networks and architectures are used to solve different tasks. For instance, \acp{CNN}, which are commonly used in computer vision and image classification applications, can recognise characteristics and patterns within an image, allowing tasks such as object detection and recognition to be accomplished. However, in tasks with a different nature, such as natural language processing, speech recognition, or timeseries forecasting \citep{Jahanbakht2021a}, Recurrent Neural Networks (RNNs) are commonly employed.
Despite the differences in their architectures, many \ac{DL} techniques, use the concept of supervised learning to process their input data and accomplish different tasks.

%The above describes the most basic form of \ac{DNN}. \ac{DL} techniques, on the other hand, are extremely complicated, and different types of neural networks are used to solve certain tasks.As an example,\ac{CNN}, which are commonly used in computer vision and image classification applications, can recognise characteristics and patterns within an image, allowing tasks such as object detection and recognition to be accomplished. Because it uses sequential or time-series data, Recurrent Neural Networks (RNNs) are commonly employed in natural language and speech recognition applications.

 %\input{tables/table_imagenet}
%  \ref{table:imagenet}

% \begin{figure}[!t]
% \centering
% 	\begin{neuralnetwork}[height=5]
% 		\newcommand{\nodetextclear}[2]{}
% 		\newcommand{\nodetextx}[2]{$x_#2$}
% 		\newcommand{\nodetexty}[2]{$y_#2$}
% 		\inputlayer[count=4, bias=false, title=Input\\layer, text=\nodetextx]
% 		\hiddenlayer[count=5, bias=false, title=Hidden\\layer, text=\nodetextclear] \linklayers
% 		\outputlayer[count=2, title=Output\\layer, text=\nodetexty] \linklayers
% 	\end{neuralnetwork}
% 	 \caption{A diagram of a single-layer neural network, composed of input, hidden, and output layers.}
% \label{fig:nn}
% \end{figure}

%%%%%%%%%%%%%%%%%%%%%%%%%%%%%%%
\subsection{Supervised Learning}

Supervised learning is a method used to enable finding and optimising a function that maps an input to its corresponding output in an input-output object pair, also known as training example  \citep{Kotsiantis2007}. 
Supervised learning uses a set of training examples based on manually-labelled training data prepared by human observers or 'supervisors', hence the name for the learning method.

The aim of supervised learning is to generate an inferred function, $f$, that maps to the training examples, and can then be used to map to new examples outside of the training examples.
In order to accomplish any general task, a computer can be programmed to find function $f$ to map $X$ to $Y$, \textit{i.e.} ($f: X \mapsto Y$), where $X$ is an input domain and $Y$ is an output domain.
For example, in an image classification task, $X$ is the dataset of images and $Y$ is a set of corresponding classification labels, which determine whether an object is present in the respective image in the dataset or not.

%Unfortunately, most of the time, it is difficult to manually feature engineer (i.e implementing domain knowledge of a dataset to create 'features' or reduced dimensionalization of the pixel values that allows  \ac{DL} algorithms to be more efficient) in order 
 
To determine the function $f$ that can recognise, for instance, a fish in an image using \ac{DL}, one solution is to do feature engineering. However, it is usually very difficult to perform this, \textit{i.e.} hand-pick features of the fish, based on the domain knowledge that comes from the training dataset. In addition, most of the time, the hand-picked features need to be pruned to reduce their pixel dimensionality.   
Comparatively, it is often more feasible to collect a large dataset of $(x, y) \in X \times Y$ to find the mapping function $f$, and this affords supervised learning  advantage as an alternative mapping technique compared with direct feature engineering.
Specifically, in the fish classification task, a large dataset of fish images is collected, where each image $x$ is labelled with $y$ that shows the presence or absence of a fish, without the need to hand-pick its features.

% When conducting supervised learning on a dataset of $n$ example images $\left\{\left(x_{1}, y_{1}\right), \ldots\left(x_{n}, y_{n}\right)\right\}$, the mapping function ($f: X \mapsto Y$) can be identified from a set of functions by searching and selecting the function that is most suitable relative to the other functions within the training dataset.
% More specifically, consider a class of functions $\mathcal{F}$ that map $X \mapsto Y$, and a scalar-valued loss (error) function $L(\hat{y}, y)$
%  that computes the difference between the predicted label $\hat{y}_{i}=f\left(x_{i}\right)$ for 
% $f \in \mathcal{F}$ and the true label $y_{i}$ to find a mapping function $f^{*}$ that minimizes the loss over the training dataset.
% Upon finding this function, the original training dataset can then be discarded but the learned function $f^{*}$ is retained, which is used to map new elements of $X$ to $Y$.

One of the main supervised learning approaches is training a neural network, which is the foundation of deep learning, especially for computer vision applications such as fish image processing. We, therefore, dedicate the next subsection to neural networks and their underlying working principles.

% \input{tables/box3}
%%%%%%%%%%%%

\subsection{Neural Networks}

A 'neural network'  \citep{Cook2020} is a computer program originally conceived by mimicking actual cerebral neural networks that make up the brain's grey matter. 
A computer's neural network, a.k.a. an artificial neural network, "learns" to do a specific task by using a large amount of data, usually through supervised network training that does not involve any task-specific rules.
As briefly mentioned, a neural network is constructed from three types of layers: an input layer, hidden or latent layers, and an output layer (see Figure \ref{fig:nn}). These layers include processing neurons within them (coloured circles in Figure \ref{fig:nn}), and connecting synapses (weights) between them (edges in the figure). 

The input layer is the gate to the network. It provides information to the network from outside data, and no calculation is made in this layer. Instead, input nodes pass the information on to the hidden layer. 
This layer is not visible to the outside world and serves as an abstraction of the inputs, independent of the neural network structure. The hidden layer (layers) processes the data received from the input layer and transfers the results to the output layer. 
Finally, the output layer brings the information that the network has learned into the outside world.

Learning in a neural network happens through minimising a loss function. Generally, a loss function is a function that returns a scalar value to represent how well the network performs a specific task. For example, in image classification, the network is expected to correctly classify all the images containing a fish as fish, and all those not including a fish, as no fish, returning a loss value of zero.  
During learning, the network receives a large amount of input data, e.g. thousands of fish images, and eventually learns to minimise the loss between its predicted output and the true target value. In the case of supervised learning, these true target values are provided to the network, to find function $f$ described in the previous section, to minimise the loss function. This minimisation happens through optimising $f$ using an algorithm such as Stochastic Gradient Descent (SGD) \citep{cosine_lr} that helps find network weights/parameters that minimise the loss.

\begin{figure*}[htbp]
\centering
\includegraphics[width=\textwidth]{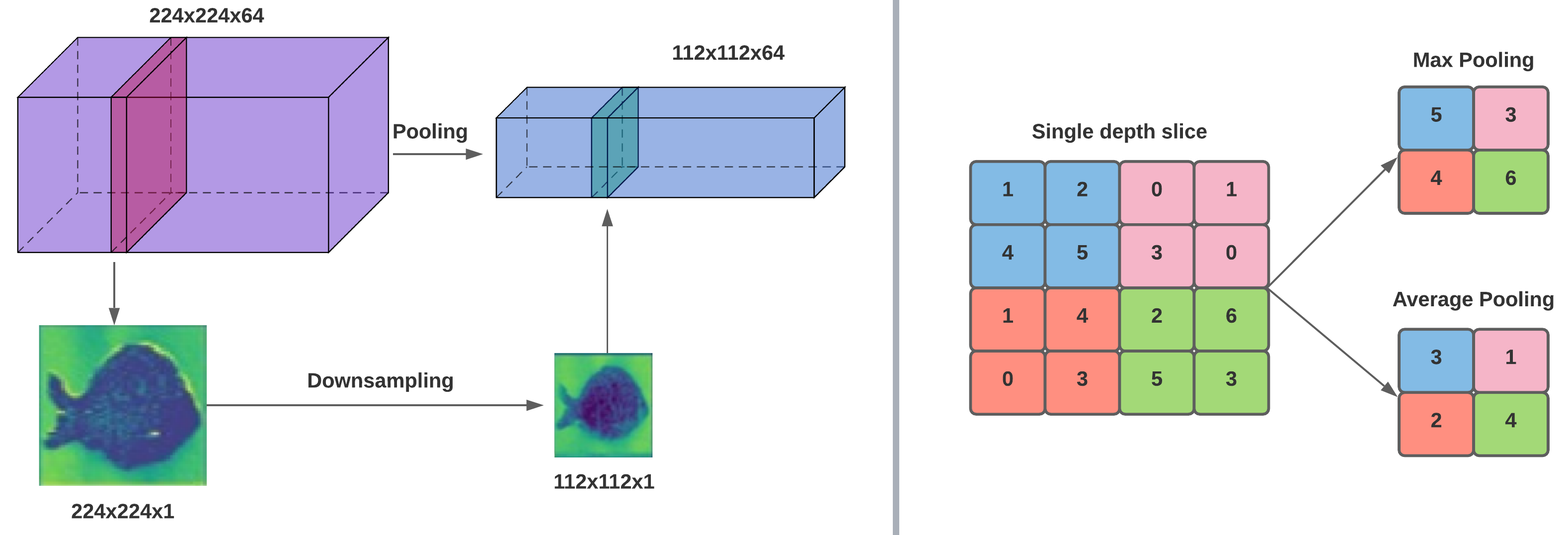}
\caption{Schematic diagram of pooling layer: (\textbf{Left}) single feature map spatially downsampled from a representation block with shape $224 \times224 \times 1 $ to a new representation of shape $112 \times 112 \times 1$.
(\textbf{Right})  types of pooling layer (max-pooling and average-pooling).}
\label{fig:pool}
\end{figure*}

\subsection{Convolutional Neural Network}
\acp{CNN} are probably the most commonly used artificial neural networks. They have been the dominant deep learning tool in computer vision and have been widely used in underwater marine habitat monitoring  \citep{Saleh2020}.
\acp{CNN} are broadly designed after the neuronal architecture of the human cortex but on much smaller scales  \citep{Schmidhuber_2015}.
A \ac{CNN}  \citep{lecun1998gradient} is specifically designed for dealing with datasets that have some spatial or topological features (e.g. images, videos), where each of the neurons are placed in such a manner that they overlap and thus react to multiple spots in the visual field. %, and each spot has redundant neurons connections. 
A \ac{CNN} neuron is a simple mathematical design of the human brain's neuron that is utilised to transform nonlinear relationships between inputs and outputs in parallel. %(see Figure \ref{fig:nn}).
There are two primary layer types in a \ac{CNN}, i.e. convolutional layers and pooling layers, which generate feature maps, as explained in the following subsections.

% \begin{figure*}[htbp]
% \centering
% % \includegraphics[width=0.48\textwidth]{Figuers/neuron.png}
% \includegraphics[width=\textwidth]{Figuers/neuron.png}
% \caption{\textbf{Left:} A drawing of a biological neuron and myelinated axon, with signal flow from inputs at dendrites to outputs at axon terminals. 
% \textbf{Right:} A diagram of the biological inspiration behind a single \ac{CNN} neuron. Inputs $x_{i}$ interact multiplicatively with the synapses $w_{i}$. The cell body accumulates the sum of all inputs and then fires an output signal according to its activation function. If the activation function is a sigmoid non-linear function (with output range in [0,1]), then the output can be interpreted as the average firing rate of the neuron. Figure from \cite{CS231n}}
% \label{neuron}
% \end{figure*}

\subsubsection{Convolutional Layer}
In this layer, the convolutional processes (\textit{i.e.}, the multiplication of a small matrix of the input neurons by a small array of weights called filter) are used on limited fields (which depend on the size of the filter) to avoid the need to learn billions of weights (parameters), which would be required if all the neurons in one layer are connected to all the neurons in the next layer. %, as is the case in a fully connected layer. 
This excessive computation is avoided through the weight-sharing of convolutional layers combined with filters for their corresponding feature maps. %, described in subsection \ref{FeatureMaps}. 
\alz{In a convolution operation, a small matrix of the input neurons is multiplied in its same-sized matrix, called a filter. In a convolutional layer, this convolution operation happens by sliding the filter on the entire input neurons, generating a feature map. Filters work on a reduced area of the input (convolutional kernel). Convolutional layers can either use the same kernel size or they can use different kernel sizes, which makes it possible to extract complex features from the input using fewer parameters.}
In addition,  weight-sharing is useful in avoiding model overfitting, i.e. memorising the training data,  \citep{abdel2013exploring}, while also reducing computing memory requirements and enhancing learning performance  \citep{korekado2003convolutional}.

\subsubsection{Pooling Layer}
This layer is used to reduce the spatial dimension (not depth) of the input features and add control for avoiding overfitting by reducing the number of representations with a specified spatial size. Pooling operations can be done in two different ways, i.e. Max and Average pooling.
In both methods (see Figure \ref{fig:pool}), an input image is down-scaled in size, by taking the maximum of 4 pixels and down-sampling them to one pixel. %(\textit{i.e.} no trainable parameters). 
Pooling layers are systematically implemented between convolutional layers in conventional \ac{CNN} architectures.
The pooling layers work on each channel (activation map) individually and downsample them spatially.
By having fewer spatial information, pooling layers make a \ac{CNN} more computationally efficient.

\subsubsection{Feature Maps} \label{FeatureMaps}
\begin{figure*}[htbp]
\centering
\includegraphics[width=0.9\textwidth]{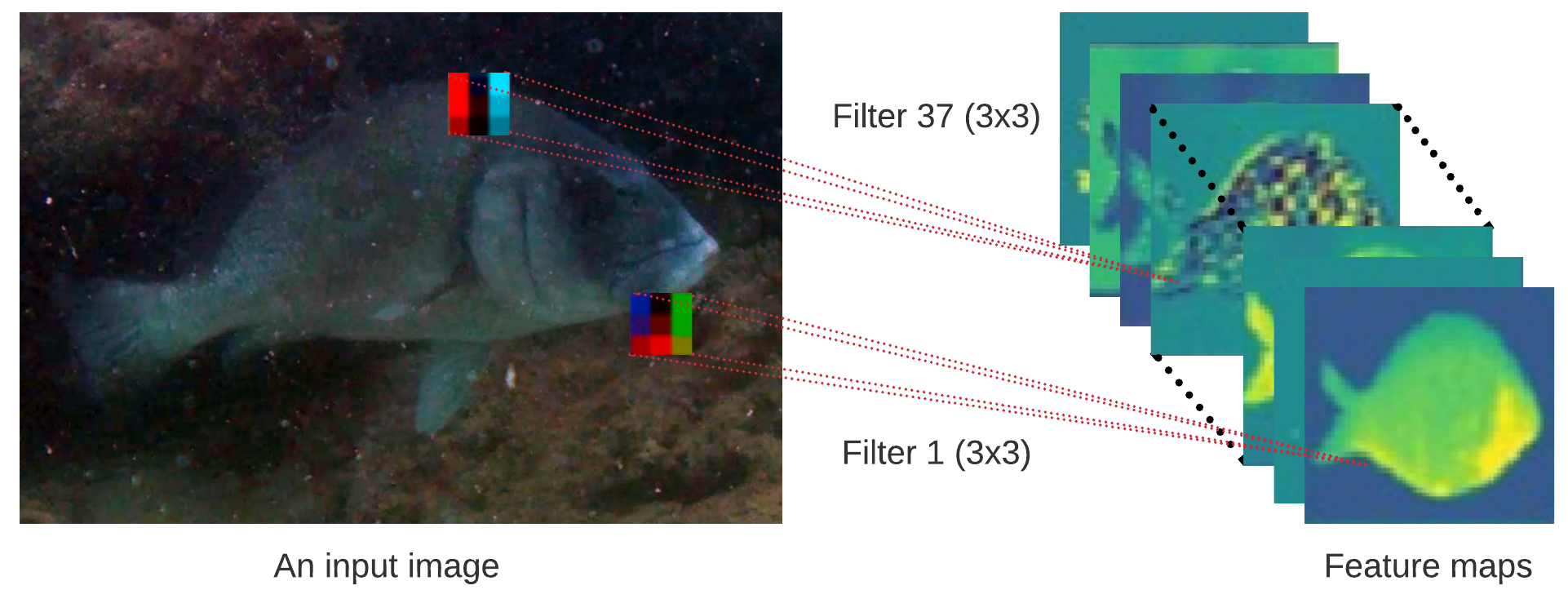}
\caption{Schematic diagram of feature maps of the \ac{CNN} used in the classification task.  The feature map is a two-dimensional representation of an input image. Here $(3 \times 3)$ is the size of the filter slid over the entire image to generate feature maps. }
\label{fig:feature}
\end{figure*}

Feature Maps, also called Activation Maps, are the result of applying convolutional filters or feature detectors to the preceding layer image. The filters are moved on the preceding layer by a specified number of pixels. %Each location activates a neuron, and the output is gathered in the feature map. 
For instance, in Figure \ref{fig:feature}, there are 37 filters of the size $3 \times 3$ that move across the input image with a stride of 1 and result in 37 feature maps. %of the size $222 \times 222$  ($224 - 3 + 1$).
% The majority of a \ac{CNN}'s layers are made up of feature maps, with each pixel acting as a neuron. 

The majority of \ac{CNN} layers are convolutional layers. These layers are used to apply the same convolutional filtering operation to different parts of the image, creating \say{neurons} that can then be used to detect features, like the edges and corners. 
A collection of weights connects each neuron in a convolutional layer to the preceding layer's feature maps, or to the input layer image. The feature maps help visualise the features that the \ac{CNN} is learning to give an understanding of the network learning process, as shown in Figure \ref{fig:feature}. 

% \acp{CNN} have been the dominant deep learning tool in computer vision and the most commonly used tool in marine and underwater fish and habitat monitoring. 
% In the next Section, we provide a comprehensive survey of how deep learning and in particular various flavours of \acp{CNN} are revolutionising marine and underwater monitoring. 

% \begin{equation*}(224 - 3 + 1)\end{equation*} 

% \section{Marine Applications}\label{sec4}

%%%%%%%%%%%%%%%%%%%%%%%%%%
%\input{tables/table_Symbols}
 
%%%%%%%%%%%%%%%%%%%%%%%%%%%%%%%%%%%%%
% \section{Applications of Deep Learning in underwater visual analysis}\label{secapps}
% \section{Applications of Deep Learning in marine science}\label{secapps}
\section{Applications of Deep Learning in Fish-Habitat Monitoring}\label{secapps}

% \alz{Although deep learning and \acp{CNN} can be used in a variety of marine applications ranging from identifying the species of harvested fish\cite{Lu2020}, to analysis of fisheries surveillance videos\cite{French2020}, to natural mortality estimation\cite{Liu2020b}, to automatic vessel detection \cite{Chen2019} and analysis of deep-sea mineral exploration
% \cite{Juliani2021},}
% Although deep learning and \acp{CNN} can be used in a variety of marine applications such as automatic vessel detection \cite{Chen2019}
% or for analysis of deep-sea mineral exploration
% \cite{Juliani2021}, 
% (XXX Give some examples other than the one we cover below, and cite a few papers on \ac{DL} for other things in marine), 
% in this paper, we focus on using \acp{CNN} for \ac{CV} tasks. 
\alz{In a recent special issue titled "Applications of machine learning and artificial intelligence in marine science" published in the International Council for the Exploration of the Sea (ICES) journal of marine science  \citep{Proud2020}, many uses of  deep learning and \acp{CNN} have been shown. These include identifying the species of harvested fish \citep{Lu2020}, analysis of fisheries surveillance videos \citep{French2020}, and natural mortality estimation \citep{Liu2020b}. Other published works have used CNN for other marine applications such as automatic vessel detection \citep{Chen2019}, and analysis of deep-sea mineral exploration \citep{Juliani2021}. However, in this paper we focus on using CNNs for CV tasks. }

These tasks are mainly designed to extract knowledge from underwater videos and images. Despite the recent use of \acp{CNN} for various visual analysis tasks such as segmentation  \citep{Garcia,Alshdaifat2020,Islam2020,Zhang2021}, localisation  \citep{SuY2020,Jalal2020,Knausgard2021a}, and counting  \citep{Tarling2021DEEPIMAGES,Schneider2020CountingLearning,Ditria2021a}, the most common and the widest studied \ac{CV} task in underwater fish habitat monitoring has been classification. Therefore, in this paper, we focus mainly on classification of underwater fish images. We survey some of the latest works on fish classification and provide a high-level technical discussion of these works.
%four of the most widely used \ac{CV} tasks facilitated by \ac{DL} for marine applications.
%These include classification, counting, localisation, and segmentation for underwater visual analysis.

%Here, the goal is to assist the readers in understanding the similarities and differences across a wide range of \ac{CV} and \ac{DL} techniques. 
%The suggested four-category taxonomy provides a framework for researchers to comprehend existing research and highlights challenges and perspective for future research.
%The rest of this section is structured into four sub-sections for each \ac{CV} task. The details of current underwater fish visual analysis using \ac{DL} are presented in  Tables \ref{table:cls}, \ref{table:cont}, \ref{table:loc}, and \ref{table:seg}.
%To provide a background on the terminology that will be consistently used in this Section, Table \ref{table:symb} describes the performance metrics, the symbol used for each metric, and a short description of what that metric represents. 

% \ref{table:cls}
% \ref{table:imagenet}
% \ref{table:cont}
% \ref{table:dataset}
% \ref{table:loc}
% \ref{table:seg}
% \ref{table:symb}

%\subsection{Classification}
% \label{seccls}

The task of classification is defined as classifying the input samples into different categories, usually based on the presence or absence of a certain object/class, in binary classification; or the presence of several different objects belonging to different classes, in multi-class classification  \citep{IsmailFawaz2019}.
Similarly, image classification is concerned with assigning a label to a whole image based on the objects in that image. Conceivably, an image can be labelled as fish, when there is a fish present in it, or negative when no fish is present. 
Similarly, images of different species should be automatically assigned to their respective classes or given a label representing their class.

Classification is a difficult process if done manually, because an image may need to be categorised into more than one class. In addition, there may be thousands of images to be classified, which makes the task very time-consuming and prone to human error. Consequently, automation can help perform classification quicker and more efficiently. 
%Consider a manual procedure in which images are compared and relative ones are classified based on similar features, but without necessarily knowing what you are searching for in advance. 
%This is a difficult assignment as the number of images in the dataset could be in thousands. 
%Moreover, many image classification tasks involve images of different objects. 
%It rapidly becomes clear that an automatic system is required to complete this task quickly and efficiently.

In the context of fish and marine habitat monitoring, \ac{CV} offers a low-cost, long-term, and non-destructive observation opportunity. One of the initial tasks performed using deep learning on \ac{CV}-collected marine habitat images is fish classification, which is a key component of any intelligent fish monitoring systems, because it may activate further processing on the fish image. % may give data support for marine researchers. 
However, underwater monitoring based on image and video processing pose numerous challenges related to the hostile condition under which the fish images are collected. These include poor underwater image quality due to low light and water turbidity, which result in low resolution and contrast.
Additionally, fish movements in an uncontrolled environment can create distortion, deformations, occlusion, and overlapping.
Many previous works \citep{Boom2012, Takada2014, Martinez-DeDios2003} have tried to address these challenges. Some of these works focused on devising new methods to properly extract traditional low-level features such as colours and textures using mean shift algorithm \citep{Boudhane2016},
in the presence of the challenges. However, these works have not been very successful compared to \ac{DL} approaches.   

With the inception of \acp{CNN}, many researchers utilised them to extract both high-level and low-level features of input images. These features, which can be automatically detected by the CNN, carry extensive semantic information that can be applied to recognise objects in an image. In addition, \acp{CNN} have the ability to address the challenges outlined above. Therefore, they are currently the main underwater image processing tool in literature for fish classification, as shown in Tables \ref{table:cls1} and \ref{table:cls2}. These tables list some of the latest classification works, while providing details about the \ac{DL} models used and the framework within which the model was implemented. It also provides information about the data source, as well as the pre-processing of the data and its labels, while reporting the \ac{CA} %(see Table \ref{table:symb}) 
and a short comparison with other methods if the reviewed work has provided it. %We hope that this Table provides a quick and easy way of assessing various underwater classification methodologies and tasks, which can be helpful in marine research.
One of the main metrics when comparing different methods for classification is their CA, which is defined as the percentage of correct predictions by the network.
\begin{equation}
 CA = (TP + TN) / (TP + TN + FP + FN),
\end{equation}
where TP (True Positive) and TN (True Negative)  represent the number of correctly classified instances, while FP (False Positive) and FN (False Negative) represent the number of incorrectly classified instances.  
For multi-class classification, CA is averaged among all the classes.

\alz{
\ac{DL} algorithms are gaining momentum in their growing accuracy in different applications. However, they have inherent limitations, which should be considered before choosing a DL algorithm for a given application. This is  because accuracy, for example in a fish classification task, may significantly differ from true accuracy due to the distribution of samples in the training and testing populations. To address this limitation of classification accuracy, the Receiver Operating Characteristics (ROC) \citep{Krupinski2017} and Area Under The Curve (AUC) \citep{Janssens2020} are widely used as a standard measure for determining the performance of a model in a binary classification setting. Their definition is very similar to  accuracy but they help one understand the probability that the classifier produces correct outputs with desired levels of true positives and false negatives,  using a certain classification threshold. 
}

%The details of underwater fish classification using \ac{DL} are shown in  
The works in Tables \ref{table:cls1} and \ref{table:cls2} can be divided into two general categories. The first category deals with designing effective \acp{CNN} that address the challenge of unconstrained, complex, and noisy underwater scenes, while the second category also tries to address the usual problem of limited fish training datasets.

As mentioned, when processing unconstrained underwater scenes specific attention should be paid to implementing a classification approach that is capable of handling variations in light intensity, fish orientation, and background environments, and similarity in shape and patterns among fish of various species.
%However, based on the literature, unconstrained underwater scenes and limited datasets are the two main challenges to do automatic classification using image processing techniques that accurately identify between classes or species of fish.
%The first challenge is the nature of unconstrained underwater scenes as it is very unstable owing to variations in light intensity, fish orientation, a diversity of background environments, and similarity in shape and pattern among fish of various species.
%The problem of recognition and automatic classification of fish in uncontrolled video sequences is not only of great interest in industry but also of great importance in the field of marine biology in general.
In order to overcome these challenge and to improve classification accuracy, various works have devised different methodologies. In  \citep{Varalakshmi2019a}, the authors used different activation functions to examine the most suitable for fish classification, while in   \citep{Sarigul2017c} different number of convolutional layers and different filter sizes were examined. 
In  \citep{Salman2016}, the authors used a \ac{CNN} model in a hierarchical feature combination setup to learn species-dependent visual features for better accuracy. In another work  \citep{Qin2016a}, principal-component analysis was used in two convolutional layers, followed by binary hashing in the non-linear layer and block-wise histograms in the feature pooling layer.
Furthermore, a single-image super-resolution method was used in  \citep{Sun2017a} to resolve the problem of limited discriminative information of low-resolution images.
Moreover,  \citep{Chen2018c} used two independent classification branches, with the first branch aiming to handle the variation of pose and scale
of fish and extract discriminative features, and the second branch making use of context information to accurately infer the type of fish.
The reviewed works show that depending on the type of environment and fish species similarities in the dataset under consideration, various techniques should be considered and investigated to find the best classification accuracy.

\alz{
As already mentioned, data gathering in the wild is sometimes very difficult and challenging, thus to maximize the success rate of training, it is essential to consider gathering field data from the beginning of the project. This ensures that the collected training dataset has good sample diversity including samples collected at different environmental conditions such as water turbidity and salinity, and it captures fish species similarities. Diversity and comprehensiveness in the dataset is one of the key factors in reaching high  classification accuracies when the model is deployed in the real world. Data augmentation is another important method that can help improve the classification accuracy, through increasing the dataset size and diversity. An alternative to data augmentation is transfer learning, but the model should be always fine-tuned to the new dataset to maximize accuracy. Image pre-processing is another important technique that can help improve classification accuracy, and should be considered when working with new fish datasets.
}

Dataset limitation, i.e. having limited number of fish images from different species, and/or having few numbers of different fish etc, is another challenge in underwater fish habitat monitoring in general and in fish classification, in specific. This challenge has been addressed in 
 \citep{Saleh2020, Jin2017a, Rathi_2017, Tamou2018a} using transfer learning.

%Dataset limitations in \ac{DL} is one of the key issues that prevent the wide adoption of \ac{DL} in a field such as a computer vision. Currently, \ac{DL} works is limited by the data, and in particular, the data size needed for training a model. In the field of \ac{CV}, only a few datasets of large size are available \cite{ILSVRC15, Lin2014, Everingham2010}, and thus there are few large-scale models that are able to generate reasonable results. On the other hand, in the field of marine science, the availability of large-scale labelled images corpus is a big issue, i.e. having a limited number of fish images from different species, and/or having few numbers of different fish etc, is another challenge in underwater fish habitat monitoring in general and in fish classification, in specific.
%Some recent works \cite{Saleh2020, Jin2017a, Rathi_2017, Tamou2018a} demonstrate that the data size limit for \ac{DL} can be relaxed to around one order of magnitude by using transfer learning.
Transfer learning is a \ac{ML} method that works by transferring information obtained while learning one problem or domain to a different but related problem or domain. 
Comparing a randomly initialised classifier with another one pre-trained on ImageNet  \citep{ILSVRC15}, Saleh \textit{et al.}  \citep{Saleh2020} achieved a fish classification accuracy of $99\%$, outperforming the randomly-initialised classifier, significantly.  
This finding shows that transfer learning can bring learned information from the ImageNet learning domain to fish classification domain and can be a useful and crucial method for evaluating fish environments.
Transfer learning was also used in  \citep{Konovalov2019UnderwaterSupervisionb}
where general-domain above-water fish image learning was transfered and used for underwater fish classification. 
In the same way, to train large-scale models that are able to generate reasonable results,  \citep{Zhuang2020} collected 1000 fish categories with 54,459 unconstrained images from various professional fish websites and Google engine.

%Table \ref{table:cls} also shows the usefulness of transfer learning in fish classification. 
In addition to transfer learning, some works have developed specific machine learning techniques suiting their applications. For instance, in a previous study  \citep{Siddiqui2018a},  %\ac{CNN}'s are capable of extracting features from images of fish.
a pre-trained \ac{CNN} was used as a generalised feature extractor to avoid the need for a large amount of training data. The authors showed that by feeding the CNN-extracted features to a Support Vector Machine (SVM) classifier \citep{Pisner2019}, a \ac{CA} of $94.3\%$ for fish species classification can be achieved, which significantly outperforms a stand-alone \ac{CNN} achieving an accuracy of 53.5\%. Also,  \citep{Deep2019a} used the same techniques in  \citep{Siddiqui2018a} to achieve a \ac{CA} of $98.79\%$.
In addition,  \citep{Iqbal2021a} developed a new technique for fish classification by modifying AlexNet \citep{ImageNet2012} model with fewer number of layers.
Moreover,  \citep{Konovalov2019a} presented a labelling efficient method of training a CNN-based fish-detector on a small dataset by adding 27,000 above-water and underwater fish images.

\acp{CNN} are sometimes capable of surpassing human performance in identifying fish in underwater images.
By training a \ac{CNN} on $900,000$ images, Villon \textit{et al.}  \citep{Villon2018a} could achieve a \ac{CA} of $94.9\%$ while human \ac{CA} was only $89.3\%$. 
This result was achieved mainly because the \ac{CNN} was able to successfully distinguish fish that were partially occluded by corals or other fish, while human could not.
Furthermore, the best \ac{CNN} model developed in  \citep{Villon2018a} takes $0.06$ seconds on average to identify each fish using typical hardware (Titan X GPU). This demonstrates that \ac{DL} techniques can conduct accurate fish classification on underwater images cost-effectively and efficiently. This facilitates monitoring underwater fish and can advance marine studies concerned with fish ecology. 

If \ac{DL} methods are going to be deployed widely for different marine applications such as fish classification, there is a need to implement them efficiently, so that they can run on low-power embedded systems, which can run in real-time on mobile devices such as underwater drones. % to satisfy their demands requires real-time performance from \ac{DL} models.
To that end, Meng \textit{et al.}  \citep{Meng2018a} have developed an underwater drone with a panoramic camera for recognising fish species in a natural lake to help protect the environment. They have trained an efficient \ac{CNN} for fish recognition and achieved $87\%$ accuracy while requiring only $6$ seconds to identify 115 images.
This promising result shows that, \ac{DL} can be used to classify underwater fish while also satisfying the real-time conditions of mobile monitoring devices. In addition, other efficient hardware design approaches that have proven useful in reducing power consumption and increasing speed in classification task in other domains such as agriculture~ \citep{Lammie2019} can be adopted on edge underwater processors.

\alz{
In \ac{DL} applications, video storage is currently a bottleneck that may be bypassed with real-time algorithms, because they only need to store some and not all the video frames in memory and process them in-situ, as they become available. This  eliminates the time it takes for all the frames to be stored and retrieved from memory. This is helpful in situations where large amounts of data have to be processed quickly, for example, in an underwater fish observation camera, where frames are collected continuously and should either be stored locally or transfered to surface, which are both costly and mostly impossible. Using real-time processing algorithms, the frames are processed and only the information obtained, \textit{i.e.} the number of fish in a frame are sent or stored, which is much lighter than the entire frame.}

%For the classification of live underwater fish, \ac{DL} is sometime used to detect specific species. However, most of the time, classification is mainly used to identify whether a given object is a fish or not, where the fish species identification is not important. This is very useful in underwater habitat surveys, where the number of present fish in an image can be counted. In the next subsection, we review previous works on counting fish in underwater images.   
%\ac{DL} can be a realistic \ac{CV} solution nowadays, where vast volumes of visual data can be acquired quickly. 
%As a result, it is worthwhile to investigate the performance levels that may be obtained by integrating \ac{DL} with \ac{CV} in order to investigate quick and accurate systems.

% The primary drawback of \ac{DL} is that it needs a huge quantity of annotated training data, and collecting and annotating large numbers of images/videos is time-consuming and tedious. 
% Section \ref{secdchl} suggest some solution for this challenge.

% \input{tables/table_cls}
% \usepackage{array}
% \usepackage{graphicx}
% \usepackage{colortbl}

% \begin{center}
% \begin{longtable}
\begin{table*}
\begin{sideways}
\begin{minipage}{\textheight}

\centering
\caption{Classification}
\label{table:cls1}
\arrayrulecolor[rgb]{0.647,0.647,0.647}
\resizebox{\linewidth}{!}{%
\begin{tabular}{>{\hspace{0pt}}p{0.18\linewidth}>{\hspace{0pt}}p{0.05\linewidth}>{\hspace{0pt}}p{0.044\linewidth}>{\hspace{0pt}}p{0.16\linewidth}>{\hspace{0pt}}p{0.16\linewidth}>{\hspace{0pt}}p{0.100\linewidth}>{\hspace{0pt}}p{0.048\linewidth}>{\hspace{0pt}}p{0.069\linewidth}>{\hspace{0pt}}p{0.186\linewidth}}
\arrayrulecolor{black}\hline
\multicolumn{1}{>{\centering\hspace{0pt}}m{0.18\linewidth}}{\textbf{Article }} & \multicolumn{1}{>{\centering\hspace{0pt}}m{0.05\linewidth}}{\textbf{DL Model}} & \multicolumn{1}{>{\centering\hspace{0pt}}m{0.044\linewidth}}{\textbf{Framework}} & \multicolumn{1}{>{\centering\hspace{0pt}}m{0.16\linewidth}}{\textbf{Data}} & \multicolumn{1}{>{\centering\hspace{0pt}}m{0.16\linewidth}}{\textbf{Annotation/Pre-processing/Augmentation}} & \multicolumn{1}{>{\centering\hspace{0pt}}m{0.100\linewidth}}{\textbf{Classes and Labels}} & \multicolumn{1}{>{\centering\hspace{0pt}}m{0.048\linewidth}}{\textbf{Perf. Metric}} & \multicolumn{1}{>{\centering\hspace{0pt}}m{0.069\linewidth}}{\textbf{Metric Value}} & \multicolumn{1}{>{\centering\arraybackslash\hspace{0pt}}m{0.186\linewidth}}{\textbf{Comparisons with other methods}} \\
\arrayrulecolor[rgb]{0.647,0.647,0.647}\hline

   \rowcolor[rgb]{0.929,0.929,0.929}  Recognition of Fish Categories Using Deep Learning Technique  \citep{Varalakshmi2019a}& CNN & Keras, Tensorflow &
  Authors-created dataset containing
  ~560 fish images,
  400 training and 160 test images. & Each image is assigned the fish species name as a label & 10 classes of 10 different fish species & CA & 95\% & NA \\
  
    Comparison of Different DL Structures for Fish Classification  \citep{Sarigul2017c} & CNN & Torch & The public QUT fish dataset 
  contains 3960 images of 468 fish species in different environments. & Each image is assigned the fish species name as a label & 468 classes of 468 different fish species & CA & 46.02\% & NA \\
  
    \rowcolor[rgb]{0.929,0.929,0.929}  Fish Species Classification in Unconstrained Underwater
  Environments Based on DL  \citep{Salman2016} & CNN & NA & The images are from the public Fish4Knowledge dataset (LifeCLEF 2014, LifeCLEF
  2015) & Each image is assigned the fish species name as a label & 25 classes of 25 different fish species & CA & 96.75\% & Comparison with the conventional SVM machine learning tool that achieved 83.94\% \\
 
   Deep-Fish: Accurate Underwater Live Fish Recognition with a DL Architecture  \citep{Qin2016a} & CNN & Matlab & The images are from the public Fish4Knowledge dataset & Each image is assigned the fish species name as a label & 23 classes of 23 Different
  fish species & CA & 98.64\% & Comparison with conventional machine learning tools as baseline methods achieving 93.58\% \\
  
   \rowcolor[rgb]{0.929,0.929,0.929} Fish Recognition from Low-resolution Underwater Images  \citep{Sun2017a} & CNN & NA & 93 videos from LifeCLEF 2015 fish dataset & Each image was annotated by drawing a bounding box and labelling by species name & 15 classes of 15 different fish species & CA & 76.57\% & Authors used the
  traditional gabor features and dense sift
  features that generated CA of 38.28\% and 28.63\%, respectively. \\
  
  Automatic Fish Classification System
  Using DL  \citep{Chen2018c} & CNN & NA & Eight target categories: Albacore tuna, Bigeye tuna, Yellowfin tuna, Mahi Mahi, Opah, Sharks, Other. & Each image is assigned the fish species name as a label & 8 classes of 8 different
  fish species & CE & 0.578, 
  1.387 & Ranked 17th on Kaggle leaderboard on test
  set at stage 1 and 16th at stage 2. \\
  
   \rowcolor[rgb]{0.929,0.929,0.929} A Realistic Fish-habitat Dataset to Evaluate DL Algorithms
  For Underwater Visual Analysis  \citep{Saleh2020}
  & ResNet-50 CNN & Pytorch & Authors-created database containing
  39,766 images from 20 habitats in remote coastal
  marine environments of tropical Australia & point-level and semantic segmentation labels & 20 classes of 20 different
  fish species & CA & 0.99 & NA \\

   Deep Learning for Underwater Image Recognition in Small Sample Size Situations  \citep{Jin2017a} & CNN & Caffe & The images are from the public Fish4Knowledge dataset & Each image is assigned the fish species name as a label & 10 classes of 10 different fish species & CA & 85.08\% & NA \\
 
   \rowcolor[rgb]{0.929,0.929,0.929}  Underwater Fish Species Classification using CNN and DL  \citep{Rathi_2017} & CNN & NA & 27000 images from the public Fish4Knowledge dataset & Each image is assigned the fish species name as a label & 23 fish classes & CA & 96.29\% & NA \\
 
\hline
\end{tabular}
}
\arrayrulecolor{black}

\end{minipage}
\end{sideways}
\end{table*}
% \end{longtable}
% \end{center}
% \usepackage{array}
% \usepackage{graphicx}
% \usepackage{colortbl}

% \begin{center}
% \begin{longtable}
\begin{table*}
\begin{sideways}
\begin{minipage}{\textheight}

\centering
\caption{Classification}
\label{table:cls2}
\arrayrulecolor[rgb]{0.647,0.647,0.647}
\resizebox{\linewidth}{!}{%
\begin{tabular}{>{\hspace{0pt}}p{0.18\linewidth}>{\hspace{0pt}}p{0.05\linewidth}>{\hspace{0pt}}p{0.044\linewidth}>{\hspace{0pt}}p{0.16\linewidth}>{\hspace{0pt}}p{0.16\linewidth}>{\hspace{0pt}}p{0.100\linewidth}>{\hspace{0pt}}p{0.048\linewidth}>{\hspace{0pt}}p{0.069\linewidth}>{\hspace{0pt}}p{0.186\linewidth}}
\arrayrulecolor{black}\hline
\multicolumn{1}{>{\centering\hspace{0pt}}m{0.18\linewidth}}{\textbf{Article}} & \multicolumn{1}{>{\centering\hspace{0pt}}m{0.05\linewidth}}{\textbf{DL Model}} & \multicolumn{1}{>{\centering\hspace{0pt}}m{0.044\linewidth}}{\textbf{Framework}} & \multicolumn{1}{>{\centering\hspace{0pt}}m{0.16\linewidth}}{\textbf{Data}} & \multicolumn{1}{>{\centering\hspace{0pt}}m{0.16\linewidth}}{\textbf{Annotation/Pre-processing/Augmentation}} & \multicolumn{1}{>{\centering\hspace{0pt}}m{0.100\linewidth}}{\textbf{Classes and Labels}} & \multicolumn{1}{>{\centering\hspace{0pt}}m{0.048\linewidth}}{\textbf{Perf. Metric}} & \multicolumn{1}{>{\centering\hspace{0pt}}m{0.069\linewidth}}{\textbf{Matric Value}} & \multicolumn{1}{>{\centering\arraybackslash\hspace{0pt}}m{0.15\linewidth}}{\textbf{Comparisons with other methods}} \\
\arrayrulecolor[rgb]{0.647,0.647,0.647}\hline

  Underwater Live Fish Recognition
  by Deep Learning  \citep{Tamou2018a} & AlexNet CNN & Matlab & 27000 images from the public Fish4Knowledge dataset 
  & Each image is assigned the fish species name as a label & 23 classes of 23 different fish species & CA & 99.45\% & NA \\

  \rowcolor[rgb]{0.929,0.929,0.929}  WildFish++: A Comprehensive Fish Benchmark for Multimedia Research  \citep{Zhuang2020} & CNN & NA & Authors-created dataset of 54,459 labelled images from various professional websites and
  Google engine & Each image is assigned the fish species name as a label & 100 classes of 1000 different fish species & CA & 74.7\% & Comparison with other state-of-the-art approaches \\

 Automatic Fish Species Classification in Underwater Videos: Exploiting Pre-trained DNN Models to
 Compensate for Limited Labelled Data  \citep{Siddiqui2018a} & CNN & MATLAB & The dataset contains 50 to 120 10-second video clips of 16 species from Western Australia during 2011 to 2013. & Each image is assigned the fish species name as a label & 16 classes of 16 Different
  fish species & CA & 89.0\% & Comparison of their proposed method of CNN+SVM achieving a CA of 89.0\% with two previous works; 
  SRC (Hsiao \textit{et al.} \citep{HSIAO201413}) 49.1\% and
  CNN (Salman \textit{et al.} \citep{Salman2016}) 53.5\%
  %Proposed method of CNN plus SVM  89.0\% 
  \\

  \rowcolor[rgb]{0.929,0.929,0.929}  Underwater Fish Species Recognition using Deep
  Learning Techniques  \citep{Deep2019a} & CNN & Keras, Tensorflow & 35000 images from the public Fish4Knowledge dataset & Each image is assigned the fish species name as a label & 23 classes of 23 different fish species & CA & 98.79\% & NA \\
  
  Automatic Fish Species Classification Using Deep
  Convolutional Neural Networks  \citep{Iqbal2021a} & modified AlexNet
  CNN & Tensorflow & The images are from two public datasets: QUT fish dataset and LifeClef-15 & Each image is assigned the fish species name as a label & 6 classes of 6 different
  fish species & CA & 90.48\% & Comparing their proposed modified AlexNet achieving a CA of 90.48\% with original AlexNet CA of 86.65\% \\
  
   \rowcolor[rgb]{0.929,0.929,0.929} Underwater Fish Detection with Weak
  Multi-Domain Supervision  \citep{Konovalov2019a} & CNN & Keras, Tensorflow & Authors-created dataset of 40000 labelled fish images from video sequences & Each image is labelled as Fish or no fish & 2 classes & CA & 99.94\% & NA \\
  
  A Deep Learning Method for Accurate and Fast Identification of Coral Reef Fishes in Underwater Images  \citep{Villon2018a} & GoogLeNet CNN & Caffe & Authors-created dataset containing 450,000
  images from over 50 reef sites around the Mayotte
  island & Annotation included drawing a rectangle
  around a single fish and associating the species name as label. & 20 classes of 20 different
  fish species & CA & 94.9\% & Comparing accuracy to human experts. The rate of correct identification was 94.9\%, greater
  than the rate of correct identification by humans
  (89.3\%). \\

 \rowcolor[rgb]{0.929,0.929,0.929} Underwater-Drone With Panoramic Camera for Automatic Fish Recognition Based on Deep Learning  \citep{Meng2018a} & CNN & NA & Authors-created dataset of 100 labelled images from Google search
  engine & Each image is assigned the fish species name as a label & 4 classes of 4 different fish species. & CA & 87\% & NA \\

\hline
\end{tabular}
}
\arrayrulecolor{black}

\end{minipage}
\end{sideways}
\end{table*}

\section{Challenges and Approaches to Address them}\label{secdchl}

Despite the rapid improvement of  \ac{DL} for marine habitat monitoring through visual analysis, \alz{four} main challenges still exist. 
The first challenge is to develop models that can generalise their learning and perform well on new unseen data samples. The second challenge is limited datasets available for general \ac{DL} tasks, and in particular for marine visual processing tasks. 
The third challenge is lower image quality in underwater scenarios. \alz{The fourth challenge is the gap between \ac{DL} and ecology.}

\alz{To address these challenges, various computer algorithms and techniques have been developed. In the following subsections, we explain the challenges in detail and briefly review various approaches to address them. However, we do not intend to include details of these approaches as they are out of the scope of this paper. The interested reader is invited to refer to relevant DL materials and the cited papers.}
%In the following subsections, we discuss these challenges in more details and describe some of the approaches in literature addressing them. We believe that the reviewed approaches in addressing these common challenges can provide a quick reference for future researchers developing DL-based marine habitat monitoring tasks. 
%As already mentioned, \ac{DL} needs a vast amount of labelled training data, which is time-consuming and laborious to collect and annotate. Furthermore, the consistency of the training samples and annotations influences the \ac{DL} performance. However, preparing a large dataset with vast amount of data and its annotation is time- and cost-prohibitive. Therefore, new approaches have been proposed to utilise the smaller available datasets, instead of asking for larger datasets.
%Various approaches have been developed in  pursuit of extending \ac{DL} for application on smaller datasets and increasing models generalization performance.
% various approaches have been developed. 
% According to  the literature review, below is a brief description of these approaches to address these challenges.

\subsection{Model Generalisation} \label{secgen}
% https://journalofbigdata.springeropen.com/articles/10.1186/s40537-019-0197-0
One of the most difficult challenges in \ac{DL} is to improve deep convolutional networks generalisation abilities. This refers to the gap between a model's performance on previously observed data (\textit{i.e.} training data) and data it has never seen before (\textit{i.e.} testing data). 
A wide gap between the training and validation accuracy is usually a sign of overfitting. Overfitting occurs when the model accurately predicts the training data, mostly because it has memorised the training data instead of learning their features. 

One way to monitor overfitting is by plotting the training and validation accuracy at each epoch during training. That way, we will see that if the gap between the validation and training acuuracy/error is widening (over- or under-fitting) or narrowing (learning). A well-known and effective method for improving the generalisability of a \ac{DL} model is to use regularisation  \citep{kukavcka2017regularization}. Some of the regularisation methods applied to fish and marine habitat monitoring domains include transfer learning  \citep{Zurowietz2020}, batch normalisation  \citep{Islam2020}, dropout  \citep{Iqbal2021a}, and using a regularisation term  \citep{Tarling2021DEEPIMAGES}.

\subsection{Dataset Limitation} \label{seclimit}
Another challenge of training \ac{DL} models is the limited dataset. \ac{DL} models require enormous datasets for training. Unfortunately, most datasets are large, expensive, and time-consuming to build. For this reason, model training is usually conducted by collecting samples from a small number of datasets, rather than from a large number of datasets. 

A dataset can be categorised into two parts: labelled data and unlabeled data. The labelled data is the set of data that needs the labelling of classes, e.g. fish species in an image, or absence or presence of fish in an image. The unlabeled data is the set of data that has not been processed. 
The labelled data forms the training set whose size is closely related to the accuracy of the trained model. The larger the training set, the more accurate the trained model. Large training set, however, are expensive to build. They require a large number of resources, such as people-hours, space, and money, making it very difficult for many researchers to achieve them, and in turn hinders their research. %This difficulty leads to an under-representation of researchers who wish to train such a model. 

%To train the deep learning model, labelled data is needed, and it is very difficult to obtain a large labelled dataset. 
Since it is difficult to obtain a large labelled dataset, various techniques have been proposed to address this challenge. %One method is to optimise the model to learn the knowledge in the unlabeled data by using the labelled data as prior knowledge. There are some methods of overcoming the data limitations such as transfer learning, data augmentation, hybrid features, and weakly supervised learning.
Some of the techniques applied to the fish and marine habitat monitoring domains include transfer learning  \citep{Qiu2018}, data augmentation  \citep{Saleh2020, Sarigul2017c}, using hybrid features  \citep{Mahmood2016,Cao2016,Blanchet2016}, weakly supervised learning  \citep{Laradji2021}, and active learning  \citep{Nilssen2017}. 

\subsection{Image Quality}
Underwater image recognition's average accuracy lags significantly behind that of terrestrial image recognition. 
This is mostly owing to the low quality of underwater photos, which frequently exhibit blurring, and colour deterioration, caused by the physical characteristics of the water and the hostile underwater environment. 

Most \ac{CV} applications perform some initial preprocessing of images before feeding them to their image processor. In underwater scenarios, these preprocessing techniques are typically used to enhance the image quality. 
Preprocessing can also help with the red channel information loss problem, which is required for obtaining relevant colour data.  
\alz{The red channel information loss problem is about losing the actual intensity of the red colour in the scene, for instance, compared to the blue and green colour channels. This is more pronounced in the underwater environment and as the depth increases, which attenuates red channel values more strongly than the other colour channels. We should, therefore, consider that the red channel value depends not only on the distance from the subject but also on the intensity of the light reflected by the subject, as the reflection of intense light is typically much stronger than that of a light of a very low intensity.}
% Another issue that arises in the remote underwater detection of a specific target in an image is the fact that multiple pixels can potentially be activated in the image. Multiple detection's of the same target within the image is possible due to, for example, a moving target, changes in the underwater environment or a combination of both.
\alz{Another issue that arises in the detection of a specific target in an underwater image is the fact that multiple pixels can potentially be activated in the image in theform of an object.  For example, sunlight shining through a periscope lens can cause spurious activation of a given pixel. There is a need for a reliable method and system for determining whether a given pixel in a remote underwater image is activated by some cause other than the presence of a target in the area of the image.}

Preprocessing of underwater photos has been extensively researched, and several solutions have been devised for correcting typical underwater image artefacts  \citep{Carlevaris-Bianco2010, Prabhakar2012}.
However, the image quality produced by these approaches is subjective to the observer, and because acquisition settings vary so widely, these methods may not be applicable to all datasets. According to empirical results \citep{Beijbom2012, Shihavuddin2013a}, the current tendency appears to be to perform picture repair and enhancement processes based on the dataset, i.e. determining the most appropriate preprocessing strategy for a specific dataset. 
\alz{This strategy also  depends on the purpose (e.g labelling, classification or both) of the images in the dataset.}
%These straightforward methods focus on dataset-specific acquisition artefacts. 
 %This implies that a target needs to be detected even if not necessarily located at a specific pixel.

In addition, basic image enhancement techniques have been shown to be effective in improving image quality. For instance, in  \citep{Cao2016} increasing  the uniformity of the background was used to boost picture contrast in underwater images for marine animal classification. 
This is a strong indicator that simple enhancing approaches might result in increased performance.
Furthermore, some recent studies have employed \ac{DL} algorithms to enhance image quality using low-quality images. 
In  \citep{He2019}, for example, end-to-end mapping is performed between low-resolution and high-resolution images. 

When compared to state-of-the-art handcrafted and traditional image enhancement methods, \ac{DL}-based algorithms typically perform better in addressing picture quality in terrestrial photos. However, significant new research is required to customise these DL-based techniques for underwater images and maritime datasets.
This poses as a future research opportunity for image quality enhancement in fish monitoring applications. Below, we discuss some more opportunities.

\subsection{Deep Learning Gap}

\alz{
\ac{DL} is an emerging field that has a lot to offer in terms of ecology. The first and most obvious ecological applications are fish classification or fish count. However, there is still a gap between the DL-predicted  fish counts and, for example, absolute abundance (fish per area or volume unit). The existing \ac{DL} literature discusses mainly the use of CNNs for the ecological problems of species classification or fish counting. However, the absolute abundance of fish is important for ecological research and species conservation.}

\alz{
Another important problem in ecological research is fish population dynamics.
A step in addressing this problem is to analyze long-term data on fish movements and fish densities. However, such long-term datasets are relatively rare and expensive to obtain. Hence, there is a need to obtain as much information as possible from the small amount of data given. This requires novel methods to give an accurate long-term estimate of fish densities or, even better, an estimate of the absolute abundance of fish.} 

\alz{
Other exemplar ecological questions that can be addressed using \ac{DL} include  species habitat selection, or the relationship between the physical environment and the life history of species  \citep{VanAllen2012,Shryock2014, Vincenzi2019}. \ac{DL} methods can help us with this because they can take advantage of all the available information. The current state of \ac{DL} research can be improved by considering alternative network architectures, more complex training algorithms, and more detailed knowledge of the problem domain. The existing \ac{DL} literature suggests that we may see many new methods in the future. Most of them still do not have sufficient data to prove that they can outperform existing methods. There are, however, examples of successful applications, such as fish classification. For many ecological problems, a \ac{DL} method can give very accurate predictions of fish densities or absolute abundance. However, it remains unclear whether this accuracy can be obtained only with the appropriate method or whether this is a property of the particular dataset on which the method was trained. From this perspective, the development of a general method for predicting fish densities and absolute abundance from very little data is a major problem in ecology. }

\alz{
One potential approach to solving this problem is to take advantage of \ac{DL} models trained on other datasets, as long as they are related to the fish density/abundance problem. The ecological literature suggests that the relationship between the physical environment and the life history of species (e.g., fish density) is likely to be complex because the physical environment differs from species to species. Therefore, we may be able to find many similar datasets on other related problems (e.g., environmental science or engineering). In addition to developing and testing general methods to estimate the absolute abundance of fish from very little data, there is a need to develop general methods that can take advantage of the ecological knowledge and domain-specific data from a particular problem. 
}

%%%%%%%%%%%%%%%%%%%%%%%%%%%%%%%%%%%%%%%%%%%%
\section{Opportunities in application of DL to fish habitat monitoring}\label{secoppt}

New methods and techniques will need to be devised to  improve the accuracy of deep learning models for various marine habitat monitoring applications and to bring them closer to their terrestrial counterparts.
% new methods and techniques should be devised. Below, we discuss some potential research opportunities, which can be useful in improving marine visual processing tasks performance and usability.  
%Devising new methods and techniques for enhancing the accuracy of underwater fish detection, counting, localisation, and segmentation from limited numbers of samples are an inevitable route for potential research due to the scarcity of datasets and the difficulty of reliable labelled data.
%Furthermore, transfer learning will help to address the issue of small sample results and the data pre-processing and augmentation will become increasingly relevant.

\subsection{Spatio-temporal and Image Data Fusion}
%\ac{CV} tasks efficiency will begin to improve as \ac{DL} models get more sophisticated and advanced. Many of the problems encountered by task-specific models can be solved by a composite model, resulting in more complex architectures. 
Most of the  current marine habitat monitoring and visual processing tools only use image-based data to train their model to understand the habitats and monitor the environment. In such tools, each frame or image is separately processed and spatiotemporal correlations across neighbouring frames are  simply overlooked. \alz{Exploiting this extra information and fusing it with the image-processing model can be beneficial  \citep{Yang2020}. 
For instance, fusing a master-slave camera setup with LSTM  \citep{Wang2017b} can help to learn the kinematic model of fish in a 3D fish tracking system.}
%To boost the overall performance, all types of \ac{DL} models as well as hand-crafted features, can be merged. 
%\ac{CNN} are commonly used, but they treat each frame separately and overlook temporal correlations across neighbouring frames.
%As a result, models that can account for spatiotemporal sequences must be considered. 
Future works should consider including spatiotemporal information in training their model and understanding the scene.
In particular, approaches similar to Long short-term memory (LSTM)  networks or other RNN models can be used in conjunction with \acp{CNN}, to obtain improved classification or prediction outcomes by taking advantage of the time-domain information. 
\alz{For example, An RNN and a CNN model are combined in  \citep{Maly2019} to achieve better performance for salmon feeding action recognition from underwater videos.  In  \citep{Peng2019}, the authors propose a spatio-temporal recurrent network to classify behavioural patterns. Similar schemes have been proposed in  \citep{Xu2021}. However, their performance and complexity heavily rely on the ability of the RNN to track the temporal relations of the frames and on the effectiveness of the CNN.}

For instance, estimating and monitoring fish development based on previous continuous observations, and analysing fish behaviour are some of the applications where time domain information will be not only useful but also critical. 
Such models can also be used to build novel video-based protocols for the surveillance of critically endangered reef fish biodiversity.

\subsection{Underwater Embedded and Edge Processing}

\acp{DNN} have proven to be successful  in both industry and research in recent years, particularly for \ac{CV} tasks. 
Specifically, large-scale \ac{DL} models have had a lot of success in real-world scenarios with large-scale data.
This is mainly due to their capacity to encode vast amounts of data and handle millions of model parameters that enhance generalisation performance when new data is evaluated.
However, this high computational complexity and substantial storage requirement makes them difficult to use in real-time applications, especially on devices with restricted resources (e.g. embedded devices and underwater edge processors for online monitoring). One approach to address this is to use compressed networks such as binarised neural networks, which have shown promise toward reaching low-power and high-speed edge inference engines  \citep{Lammie2019}, for near-underwater-sensor processing. This can significantly improve underwater image analysis capabilities, because the collected large-volume images do not need to be transferred to surface for processing, and only the low-volume results can be communicated to shore. This also solves another problem, which is the challenging underwater communication  \citep{Jahanbakht2021}. 

\subsection{Combining Data from Multiple Platforms}
%Opportunities that still exist for improving the fish monitoring using DL techniques include the combination of multiple data sets from multiple platforms to achieve a more reliable estimation of fish abundance and biomass, the use of DL methods to achieve automated measurements, the use of DL in multi-parameter modelling, the use of unsupervised DL methods, the use of DL in underwater imagery for fish monitoring, and the use of DL to address the need for large-scale monitoring.

%The combination of multiple data sets from multiple platforms to achieve a more reliable estimation of fish abundance and biomass presents an important opportunity for future research. 
The use of different data collection platforms such as autonomous underwater vehicles (AUVs) or occupied submarines, can provide different image data from different perspectives of the same or different underwater habitats, to train more effective DNNs. In addition, using simultaneous data from multiple platforms can give more monitoring information, for instance, of fish distribution patterns, especially in situations where the number of platforms is limited. 
However, combining data from multiple platforms introduces some challenges such as the lack of ground truth (e.g., the number of fish in the sampled area for all the platforms), and the need to develop techniques that can integrate these data in a robust manner. Future research can work toward addressing these challenges to exploit the significant benefits of multiple platform data combination.

\subsection{Automated Fish Measurement and Monitoring}
%In addition, other examples of opportunities that still exist for using DL methods include the use of DL to achieve automated measurements, the use of DL to address the need for large-scale monitoring, and the use of DL in underwater imagery for fish monitoring. Automated measurements can facilitate monitoring fish abundance and biomass. On the other hand, this can be achieved in scenarios such as when the monitoring locations are in remote places, when the environmental conditions may not be favourable for manual measurements, and when there is a small number of data sets. The use of DL in large-scale monitoring has several benefits, such as low computation time, which can be achieved through unsupervised methods, and the ability to model large environmental features, which can be done with semi-supervised methods. The use of DL in underwater imagery has several benefits, including the ability to obtain data from a great variety of sources, and high sensitivity to detect fish. In the future, we may expect that researchers may want to monitor biological dynamics, such as fish species presence, abundance, and biomass, along with environmental and habitat features. This may be achieved through a process that can include the combination of several approaches, including the use of DL.

DL can be used to achieve automated fish measurements, which may be useful in underwater fish monitoring, \alz{for instance to survey fish growth  \citep{Yang2020} through monitoring of fish length  \citep{Palmer2022}} and abundance \citep{Ditria2019}. \alz{Here, abundance means the number of fish in an image or video frame, and not the fish count per area or volume unit.} In addition, automated measurements can realise remote fish assessments, for example when the monitoring locations are remote, or the environmental conditions and or potential hazards do not allow frequent underwater scouting by human. 

DL can also be used for automation of monitoring of other fish biological variables such as their \alz{movement} dynamics, present species, and their abundance and biomass. On top of these, DL can be used to automate understanding of environmental and habitat features. To achieve these, new datasets should be collected, and new or existing DL techniques should be devised or customised in future research. %This may be achieved through the combination of several approaches including the DL.

\section{Conclusion}\label{seccncl}

\acf{DL} sits at the forefront of the machine learning technologies providing the processing power needed to enable underwater video to fulfill its promise as a critical tool for visual sampling of fish. It offers efficient and accurate solutions to the challenges of adverse water conditions, high similarity between fish species, cluttered backgrounds, occlusions among fish, that have limited the spatio-temporal consistency of underwater video quality. As a result, \ac{DL}, complemented by many other advances in monitoring hardware and underwater communication technologies, opens the way for underwater video to provide comprehensive fish sampling. This can span from shallow fresh and marine waters to the deep ocean, opening the way for the development of the truly comparative understanding of marine and aquatic fish fauna and ecosystems that has hitherto been impossible. At least as importantly, \ac{DL} solves the problem of handling the vast quantities of data produced by underwater video in a consistent and cost-effective way, converting a prohibitively expensive activity into a simple issue of computer processing. By enabling the processing of vast quantities of data, \ac{DL} allows underwater fish video surveys to be conducted with unprecedented levels of spatial and temporal replication enabling the massive knowledge advances that flow from the ability of underwater videos to be deployed contemporaneously across many habitats, and at many spatial scales, or to provide continuous data over time. 

\ac{DL}, and associated techniques, have the potential for widespread use in marine habitat monitoring for (1) data classification and feature extraction to improve the quality of automatic monitoring tools; or (2) to provide a reliable means of surveying fish habitats and understanding their \alz{movement} dynamics. While this will allow marine ecosystem researchers and practitioners to increase the efficiency of their monitoring efforts, effective development of \ac{DL} will require concentrated and coordinated data collection, model development, and model deployment efforts, as well as transparent and reproducible research data and tools, which help us reach our target sooner.

\section*{Acknowledgement}
This research is supported by an Australian Research Training Program (RTP) Scholarship. We acknowledge the Australian Research Council for funding awarded under their Industrial Transformation Research Program.

% \section*{acknowledgements}
% Acknowledgements should include contributions from anyone who does not meet the criteria for authorship (for example, to recognize contributions from people who provided technical help, collation of data, writing assistance, acquisition of funding, or a department chairperson who provided general support), as well as any funding or other support information.

\section*{Conflict of Interest}
All authors declare that they have no conflicts of interest.
% You may be asked to provide a conflict of interest statement during the submission process. Please check the journal's author guidelines for details on what to include in this section. Please ensure you liaise with all co-authors to confirm agreement with the final statement.

\section*{Data Availability Statement}
Data  sharing  is  not  applicable  to  this  article  as  no  data  sets  were  generated or analysed during the current study.

\section*{ORCID}
Alzayat Saleh \href{https://orcid.org/0000-0001-6973-019X}{0000-0001-6973-019X}\\
Marcus Sheaves \href{https://orcid.org/0000-0003-0662-3439}{0000-0003-0662-3439}\\
Mostafa~Rahimi~Azghadi \href{https://orcid.org/0000-0001-7975-3985}{0000-0001-7975-3985}\\

% \section*{Supporting Information}

% Supporting information is information that is not essential to the article, but provides greater depth and background. It is hosted online and appears without editing or typesetting. It may include tables, figures, videos, datasets, etc. More information can be found in the journal's author guidelines or at \url{http://www.wileyauthors.com/suppinfoFAQs}. Note: if data, scripts, or other artefacts used to generate the analyses presented in the paper are available via a publicly available data repository, authors should include a reference to the location of the material within their paper.

\printendnotes

% Submissions are not required to reflect the precise reference formatting of the journal (use of italics, bold etc.), however it is important that all key elements of each reference are included.
% \bibliography{references}

% references section
% \bibliographystyle{IEEEtran}
% \bibliographystyle{IEEEtranN}
% \bibliographystyle{WileyNJD-ACS}
% \bibliographystyle{WileyNJD-AMA}
% \bibliographystyle{WileyNJD-AMS}
% \bibliographystyle{WileyNJD-APA}
% \bibliographystyle{WileyNJD-Harvard}
% \bibliographystyle{WileyNJD-VANCOUVER}
% \bibliographystyle{apalike}
% \bibliographystyle{unsrtnat}
\bibliography{references}

% \input{tables/table_trend}
% \input{tables/table_cls_1}
% \input{tables/table_cls_2}

% % \appendix
% \clearpage
% % \vspace{15mm}
% % \section*{Appendix A}
% % \label{app:trend}
% \input{tables/table_trend}

% \clearpage

% \begin{biography}[example-image-1x1]{A.~One}
% Please check with the journal's author guidelines whether author biographies are required. They are usually only included for review-type articles, and typically require photos and brief biographies (up to 75 words) for each author.
% \bigskip
% \bigskip
% \end{biography}

% \graphicalabstract{example-image-1x1}{Please check the journal's author guildines for whether a graphical abstract, key points, new findings, or other items are required for display in the Table of Contents.}

\end{document}